%% file: acl_latex.tex
\pdfoutput=1
\documentclass[11pt]{article}

\usepackage[final]{acl}
\usepackage{latexsym}
\usepackage{wrapfig}
\input{math_commands.tex}

\usepackage{times}
\usepackage{color}
\usepackage{bm}
\usepackage{tabularx}
\usepackage{tabularray}
\usepackage{dsfont}
\usepackage[T1]{fontenc}
\usepackage[utf8]{inputenc}

\usepackage{microtype}

\usepackage{inconsolata}

\usepackage{graphicx}
\usepackage{microtype}
\usepackage{hyperref}       %
\usepackage{inconsolata}
\usepackage{algorithm}
\usepackage{wrapfig}
\usepackage{algorithmic}
\usepackage{colortbl}
\usepackage{wrapfig}
\usepackage{xcolor}
\usepackage{inconsolata}
\usepackage{multirow}  % 为了使用跨行表格加的
\usepackage{amssymb}   % 为了使用粗体R
\usepackage{amsmath}
\usepackage{color}
\usepackage{fontawesome5}
\usepackage{subfigure} % 为了使用子图
\usepackage{makecell} % 为了自定义表格长度
\usepackage{booktabs} % 为了使用三线表
\usepackage{enumitem} % 为了让项目符号取消缩进
\usepackage{multirow} % 为了让允许表格多行显示

\usepackage{booktabs} % 为了允许在合并后的内容内换行
\usepackage{stfloats}
\usepackage{listings}
\usepackage{cite}
\usepackage{pifont}
\usepackage{tcolorbox}
\usepackage{soul}
\usepackage{enumitem}
\tcbuselibrary{skins, breakable, theorems}
\usepackage{wrapfig}
\usepackage{lscape}
\usepackage{url}

\definecolor{mypink}{rgb}{.99,.91,.95}
\definecolor{myyellow}{rgb}{.99,.94,.82}

\input{config}

\title{\raisebox{-6pt}{\includegraphics[height=1.85em]{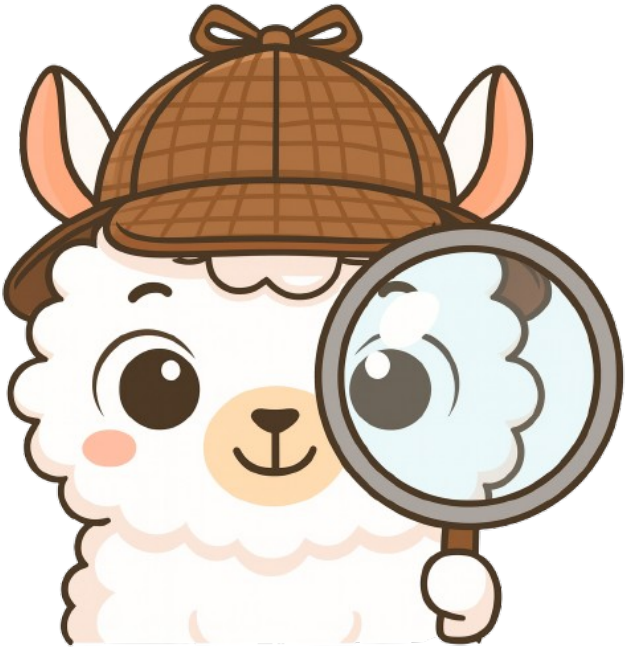}}\xspace FaithLens: Detecting and Explaining Faithfulness Hallucination}
\author{
\textbf{Shuzheng Si\thanks{\ Equal Contribution.}$^{\spadesuit\diamondsuit}$, Qingyi Wang\footnotemark[1]$^{\bigstar}$, Haozhe Zhao\footnotemark[1]$^{\clubsuit}$, Yuzhuo Bai$^{\spadesuit}$} \\
\textbf{Guanqiao Chen$^{\spadesuit}$, Kangyang Luo$^{\spadesuit}$, Gang Chen$^{\diamondsuit}$, Fanchao Qi\thanks{\ Corresponding Authors.}$^{\diamondsuit}$} \\
\textbf{Minjia Zhang$^{\clubsuit}$, Baobao Chang$^{\heartsuit}$,} and \textbf{Maosong Sun\footnotemark[2]$^{\spadesuit}$} \\ 
$^{\spadesuit}$ Tsinghua University \quad $^\diamondsuit$ DeepLang AI \quad $^{\bigstar}$ Fudan University \\
$^{\clubsuit}$ University of Illinois Urbana-Champaign \quad $^{\heartsuit}$ Peking University
}

\begin{document}

\maketitle

\renewcommand{\thefootnote}{\fnsymbol{footnote}}
\renewcommand{\thefootnote}{\arabic{footnote}}
\urlstyle{same}
\definecolor{darkgreen}{RGB}{50,100,0}
\definecolor{darkred}{RGB}{200, 0, 0}
\definecolor{lightred}{RGB}{250, 200, 200}
\definecolor{lightblue}{RGB}{210, 220, 250}
\newcommand{\cmark}{\textcolor{darkgreen}{\ding{51}}} %
\newcommand{\xmark}{\textcolor{darkred}{\ding{55}}} %
\definecolor{tabcolor1}{RGB}{247,225, 237} %lightpink
\definecolor{tabcolor2}{RGB}{255, 250, 132} %lighyellow
\definecolor{tabcolor3}{RGB}{204, 232, 207} %lightgreen
\definecolor{tabcolor4}{RGB}{245, 222, 179} %lightorange
\definecolor{tabcolor5}{RGB}{210, 220, 250} %lightblue
\definecolor{tabcolor6}{RGB}{222, 222, 222} %lightgrey

\input{Files/0_Abstract}
\input{Files/1_Introduction}
\input{Files/3_Method}
\input{Files/4_Experiment}
\input{Files/2_Related_work}

\input{Files/5_Conclusion}

% \section*{Acknowledgments}

% This document has been adapted
% by Steven Bethard, Ryan Cotterell and Rui Yan
% from the instructions for earlier ACL and NAACL proceedings, including those for
% ACL 2019 by Douwe Kiela and Ivan Vuli\'{c},
% NAACL 2019 by Stephanie Lukin and Alla Roskovskaya,
% ACL 2018 by Shay Cohen, Kevin Gimpel, and Wei Lu,
% NAACL 2018 by Margaret Mitchell and Stephanie Lukin,
% Bib\TeX{} suggestions for (NA)ACL 2017/2018 from Jason Eisner,
% ACL 2017 by Dan Gildea and Min-Yen Kan,
% NAACL 2017 by Margaret Mitchell,
% ACL 2012 by Maggie Li and Michael White,
% ACL 2010 by Jing-Shin Chang and Philipp Koehn,
% ACL 2008 by Johanna D. Moore, Simone Teufel, James Allan, and Sadaoki Furui,
% ACL 2005 by Hwee Tou Ng and Kemal Oflazer,
% ACL 2002 by Eugene Charniak and Dekang Lin,
% and earlier ACL and EACL formats written by several people, including
% John Chen, Henry S. Thompson and Donald Walker.
% Additional elements were taken from the formatting instructions of the \emph{International Joint Conference on Artificial Intelligence} and the \emph{Conference on Computer Vision and Pattern Recognition}.

% Bibliography entries for the entire Anthology, followed by custom entries
%\bibliography{anthology,custom}
% Custom bibliography entries only
\bibliography{custom}

\include{Files/6_Appendix}

\end{document}

%% file: math_commands.tex
\usepackage{amsmath,amsfonts,bm}

\def\eqref#1{equation~\ref{#1}}
\def\1{\bm{1}}

\DeclareMathAlphabet{\mathsfit}{\encodingdefault}{\sfdefault}{m}{sl}
\SetMathAlphabet{\mathsfit}{bold}{\encodingdefault}{\sfdefault}{bx}{n}

\def\gM{{\mathcal{M}}}

%% file: config.tex
\definecolor{Gray}{gray}{0.9}
\definecolor{mygreen}{rgb}{0.0, 0.5, 0.0}
\definecolor{myred}{rgb}{0.8, 0.25, 0.33}
\definecolor{myblue}{rgb}{0.19, 0.55, 0.91}
\definecolor{uclablue}{rgb}{0.15, 0.45, 0.68}
\definecolor{ucladblue}{rgb}{0.0, 0.33, 0.53}
\definecolor{ucladdblue}{rgb}{0.0, 0.23, 0.36}
\definecolor{uclagold}{rgb}{1.0, 0.82, 0.0}
\definecolor{ucladgold}{rgb}{1.0, 0.78, 0.17}
\definecolor{ucladdgold}{rgb}{1.0, 0.72, 0.11}
\definecolor{boxgreen}{rgb}{0.02, 0.66, 0.02}
\definecolor{boxred}{rgb}{0.66, 0.1, 0.1}
\definecolor{boxblue}{rgb}{0.01, 0.01, 0.73}

\usepackage{times}
\usepackage{latexsym}

\usepackage[utf8]{inputenc}
\usepackage[T1]{fontenc}
\usepackage{inconsolata}

\usepackage{algorithm}
\usepackage{algorithmic}

\usepackage{newfloat}
\usepackage{listings}
\floatstyle{ruled}
\newfloat{listing}{tb}{lst}{}
\floatname{listing}{Listing}

\usepackage{graphicx}
\usepackage{amsmath,amssymb,amsthm,mathabx,amsfonts}
\usepackage{bbm}
\usepackage{cleveref}
\usepackage{acronym}
\usepackage{enumitem}
\usepackage{balance}
\usepackage{xspace}
\usepackage{setspace}
\usepackage{url}
\usepackage{amsfonts}
\usepackage{multirow}
\usepackage{booktabs}
\usepackage{tablefootnote}
\usepackage[switch]{lineno}
\usepackage{multicol,lipsum}
\usepackage{tikz}
\usepackage{pgfplots}
\usepackage{pgfplotstable}
\usepackage{arydshln}

\usepackage{CJKutf8}

\pgfplotsset{compat=1.18}
\makeatletter
\DeclareRobustCommand\onedot{\futurelet\@let@token\@onedot}
\def\@onedot{\ifx\@let@token.\else.\null\fi\xspace}

\makeatother

\makeatletter
\newcommand{\thickhline}{%
    \noalign {\ifnum 0=`}\fi \hrule height 1pt
    \futurelet \reserved@a \@xhline
}

\crefname{algorithm}{Alg.}{Algs.}
\Crefname{algocf}{Algorithm}{Algorithms}
\crefname{section}{Sec.}{Secs.}
\Crefname{section}{Section}{Sections}
\crefname{table}{Tab.}{Tabs.}
\Crefname{table}{Table}{Tables}
\crefname{figure}{Fig.}{Fig.}
\Crefname{figure}{Figure}{Figure}
\crefname{appendix}{Appendix}{Appendices}

\acrodef{nlp}[NLP]{natural language processing}
\acrodef{plm}[PLM]{Pre-trained Language Model}
\acrodef{sota}[SOTA]{state-of-the-art}
\acrodef{icl}[ICL]{In-Context Learning}
\acrodef{bbl}[BBL]{BIG-bench Lite}

\usepackage{colortbl}
\definecolor{gblue}{HTML}{4285F4}
\definecolor{gred}{HTML}{DB4437}
\definecolor{ggreen}{HTML}{0F9D58}

\definecolor{mygray}{gray}{.92}
\definecolor{emphypurple}{rgb}{0.302, 0.055, 0.659}

\definecolor{highlightgreen}{HTML}{009901}
\definecolor{highlightred}{HTML}{FD6864}

%% file: Files/0_Abstract.tex
\begin{abstract} 
Recognizing whether outputs from large language models (LLMs) contain faithfulness hallucination is crucial for real-world applications, e.g., retrieval-augmented generation and summarization.
In this paper, we introduce \textbf{FaithLens}, a cost-efficient and effective faithfulness hallucination detection model that can jointly provide binary predictions and corresponding explanations to improve trustworthiness. 
To achieve this, we first synthesize training data with explanations via advanced LLMs and apply a well-defined data filtering strategy to ensure label correctness, explanation quality, and data diversity. 
Subsequently, we fine-tune the model on these well-curated training data as a cold start and further optimize it with rule-based reinforcement learning, using rewards for both prediction correctness and explanation quality. 
Results on 12 diverse tasks show that the 8B-parameter FaithLens outperforms advanced models such as GPT-5.2 and o3. 
Also, FaithLens can produce high-quality explanations, delivering a distinctive balance of trustworthiness, efficiency, and effectiveness. \footnote{~The data and code will be available at \url{https://github.com/S1s-Z/FaithLens}. Email: ssz24@mails.tsinghua.edu.cn.}
\end{abstract}

%% file: Files/1_Introduction.tex
\section{Introduction}
\label{section:introduction}

Recent progress in large language models (LLMs) has revolutionized text generation \citep{gpt-5}.
In practice, LLMs are widely used to generate coherent responses based on the provided contextual information, e.g., retrieval-augmented generation (RAG) \citep{wang-etal-2025-document}.
However, LLMs are prone to generating hallucinated claims that are inconsistent or irrelevant to the given context, i.e., faithfulness hallucinations \citep{ bi2024contextdpoaligninglanguagemodels, si2025teaching}.
Therefore, detecting such hallucinations is critical for providing responsible LLM services.

To identify faithfulness hallucinations in LLM-generated outputs, recent works utilize the strong generalization abilities of LLMs and formulate it as a binary classification task \citep{ wang2024halujcritiquebasedhallucinationjudge}.
The first line of research leverages designed prompts to query advanced LLMs like GPT-4o \citep{OpenAI2023GPT4TR} to check if generated outputs contain hallucinated claims \citep{liu-etal-2023-g, lei2023chainnaturallanguageinference, dhuliawala-etal-2024-chain, muhammed2025selfcheckagentzeroresourcehallucinationdetection}, e.g., SelfCheckGPT \citep{manakul-etal-2023-selfcheckgpt}.
However, these methods are inefficient for real-world deployment because they rely on large and advanced models to achieve reliable detection performance.

\begin{figure}
    \centering
    \includegraphics[width=7.8cm]{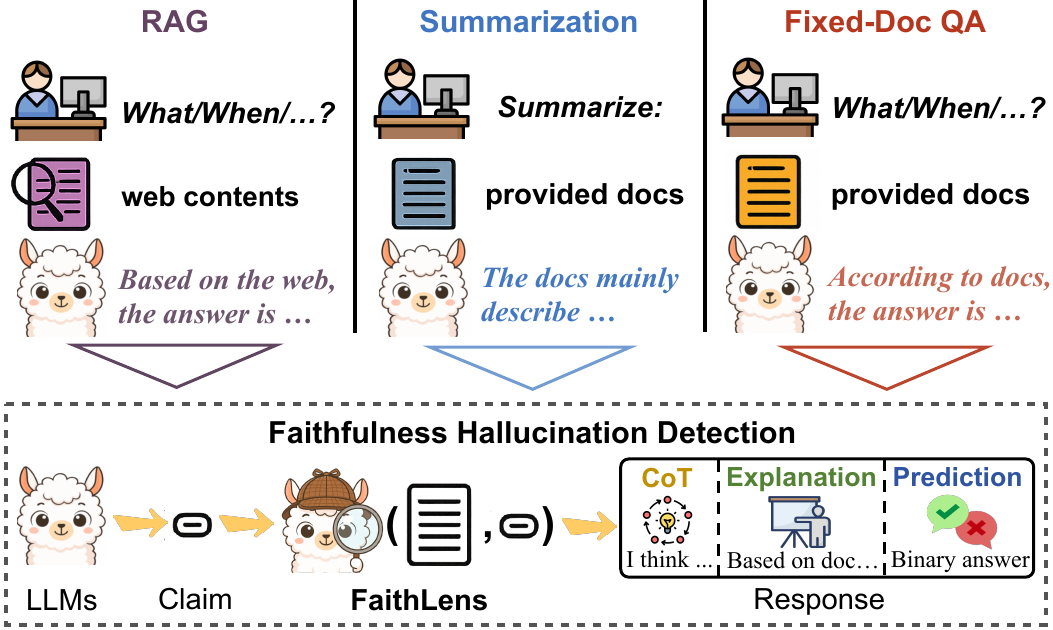}
    \caption{\textbf{The illustration of our FaithLens}. 
    Given a document $doc$ and a claim $c$, FaithLens can jointly determine whether the claim is faithful or hallucinated and provide the corresponding explanations for its decision, applicable across various tasks.}
    \label{fig_example}
\end{figure}

Thus, many studies have focused on developing cost-efficient and specialized classifiers to detect hallucinations \citep{zha-etal-2023-alignscore, seo2025verifying}.
For example, MiniCheck \citep{tang-etal-2024-minicheck} uses synthetic data generation techniques to train a 7B-parameter model, achieving performance comparable to GPT-4o.
However, developing a detection model for real-world users still faces three key challenges.
Specifically, \textbf{(1) Lack of Explainability}: Current methods typically treat faithfulness hallucination detection as a binary classification task, acting as a black box that only returns the final prediction without a corresponding explanation \citep{tang-etal-2024-minicheck}.
This makes it difficult for users to localize errors and understand why tested claims are hallucinated, which limits the trustworthiness of detection models. 
\textbf{(2) Inconsistent Generalization across Tasks}: 
Previous methods are primarily designed for detecting task-specific hallucination \citep{george-stuhlmueller-2023-factored}, e.g., summarization \citep{wan-etal-2024-acueval, hu2025divide}, and then fail to transfer across different tasks effectively.
Even the models designed for general-purpose scenarios \citep{tang-etal-2024-minicheck,lei-etal-2025-factcg, seo2025verifying} still perform unevenly on different tasks because each task may have unique hallucination patterns.
For example, summarization hallucinations typically manifest as subtly distorted content from the context \citep{li-yu-2025-summary}, whereas RAG hallucinations often ignore the retrieved context and involve conflicting claims \citep{xu-etal-2024-knowledge-conflicts}.
\textbf{(3) Lack of High-Quality Data}: Annotating training data for hallucination detection is costly and often results in low inter-annotator agreement \citep{seo2025verifying}.
Consequently, recent works propose to utilize synthetic data to train the model \citep{tang-etal-2024-minicheck, lei-etal-2025-factcg}. 
However, these methods often lack well-defined data quality control strategies.
This may result in a low-quality training set, such as ignoring data diversity and retaining too many simple instances, ultimately limiting the model's abilities in complex detection scenarios.

In this paper, we introduce a cost-efficient and effective model \textbf{FaithLens} for faithfulness hallucination detection.
As shown in Figure \ref{fig_example}, FaithLens not only predicts whether a claim is hallucinated, but also produces the corresponding explanation for users to localize errors and understand why certain claims are considered hallucinations.
To this end, we begin by leveraging open-source datasets and querying an advanced model to synthesize samples with explanations.
Next, to ensure data quality and the effectiveness of the trained model across diverse scenarios, we design a targeted data filtering pipeline that jointly ensures label correctness, the synthesized explanation quality, and data diversity. 
After using this well-curated dataset for supervised fine-tuning (SFT) as a cold start, we further strengthen the model through a rule-based reinforcement learning (RL) stage. 
Specifically, we introduce a prediction correctness reward to improve detection performance and an explanation quality reward to enhance the informativeness and clarity of generated explanations.
The correctness reward is computed directly from the model prediction, ensuring that the training signal explicitly reinforces accurate hallucination detection.
Meanwhile, our proposed explanation quality reward thoroughly assesses a generated explanation by checking if it can help a novice-level model (e.g., untuned Llama-3.1-8B-Inst \citep{llama3} model) correctly predict the corresponding label.
If the generated explanation enables a novice-level model to generate the correct prediction, it indicates that the explanation is sufficiently coherent and informative to convey the relevant evidence.
By utilizing these two rewards together with a format reward, our model can achieve a unique combination of trustworthiness and effectiveness.

We evaluate the effectiveness of our FaithLens on 12 diverse faithfulness hallucination detection tasks from LLM-AggreFact \citep{tang-etal-2024-minicheck} and HoVer \citep{jiang-etal-2020-hover}.
Experiments show that 8B-parameter FaithLens achieves state-of-the-art performance, even surpassing advanced LLMs such as GPT-5.2 \citep{gpt-5-2} and o3 \citep{jaech2024openai} with much lower cost.
Also, FaithLens can offer informative and coherent explanations, providing users with a clear understanding of why a claim is considered hallucinated. 
% To the best of our knowledge, these results are unprecedented for open-source models.

%% file: Files/3_Method.tex
\section{Task Formulation}
Given the grounding document $doc$ and the LLM-generated claim $c$, we consider $c$ to be faithful to $doc$ if a generic reader would affirm the statement ``According to the given $doc$, $c$ is true''.
Conversely, $c$ is considered hallucinated if it contradicts, misinterprets, or cannot be verified using $doc$.

Previous works \citep{laban-etal-2022-summac, zha-etal-2023-alignscore, tang-etal-2024-minicheck, lei-etal-2025-factcg,seo2025verifying} formulate such hallucination detection as a binary classification task.
The goal is to train model $\gM$ to estimate the conditional probability:
\begin{equation} 
P_{\gM}(y \mid doc, c),
\end{equation}
% where $ y \in \{0,1\}$ denotes whether the claim $c$ is faithful (1) or hallucinated (0) with respect to the given document $doc$.
where $y$ is 1 if the given claim $c$ is faithful to the document $doc$, and 0 if it is hallucinated.

In this work, we extend the standard binary classification formulation to not only predict whether a claim $c$ is faithful or hallucinated, but also provide a corresponding explanation $e$ that justifies the prediction from our model $\hat \gM$.  
Formally,
\begin{equation} 
P_{\hat \gM}(e, y \mid doc, c),
\end{equation}
where $y \in \{0,1\}$ is the prediction and $e$ is a textual explanation that support the prediction.
This formulation allows the model to provide explainable outputs that are informative to users, improving trustworthiness in hallucination detection.

\begin{figure*}
    \centering
    % \vspace{-7mm}
    \includegraphics[width=1\linewidth]{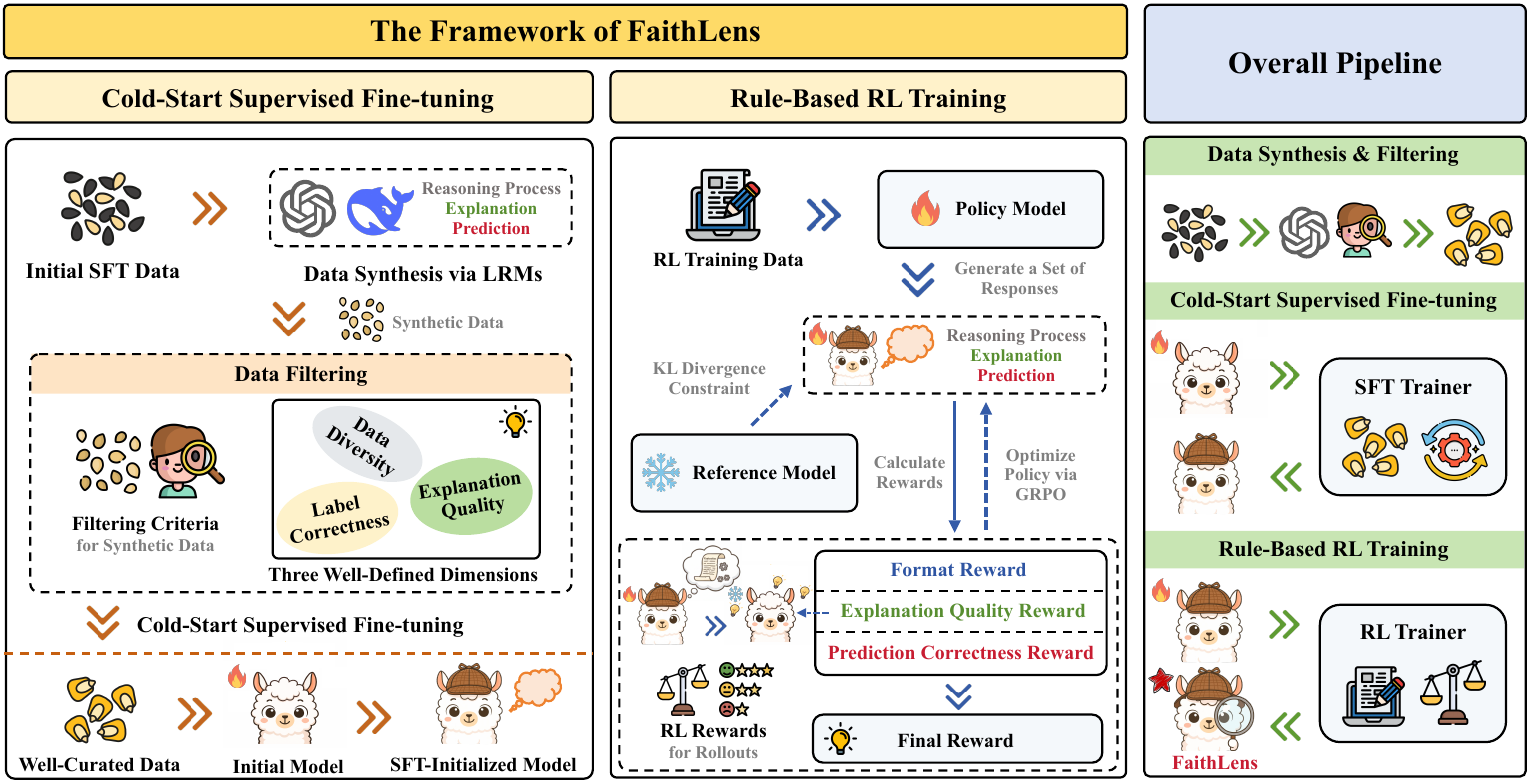}
    \caption{
     \textbf{The Overall Process of Training FaithLens}, including 
     (1) \textbf{Cold-Start SFT}: We first synthesize high-quality data with explanations used for the SFT stage.
     (2) \textbf{Rule-Based RL Training}: We further refine the model using a rule-based RL approach with the designed rewards for both prediction correctness and explanation quality.
    }
    \label{figure:model}
\end{figure*}

\section{Methodology}
\label{section:method}
In this paper, we build a cost-efficient and effective hallucination detection model \textbf{FaithLens} that can jointly determine whether the claim is faithful or hallucinated, and provide corresponding explanations to improve trustworthiness.
As shown in Figure \ref{figure:model}, we include two key stages to train FaithLens without human efforts: 
(1) A training data synthesis pipeline that first generates data with explanations, then uses a well-defined data filtering strategy to ensure data quality, and finally starts the SFT stage as a cold start (\S~\ref{sft_stage}); 
(2) A rule-based RL stage to further optimize model effectiveness and trustworthiness, using rewards from both prediction correctness and explanation quality (\S~\ref{rl_stage}).

\subsection{Cold-Start Supervised Fine-tuning}
\label{sft_stage}
To equip the model with the ability to detect hallucinations and generate corresponding explanations, we start by training the model via the SFT stage. 

\subsubsection{Data Synthesis}
\label{sec:data_syn}
Previous works \citep{tang-etal-2024-minicheck, lei-etal-2025-factcg} formulate such hallucination detection as a binary classification task and cannot provide corresponding explanations.
Thus, existing training datasets only provide prediction labels without corresponding explanations.
To bridge this gap, we first leverage the advanced large reasoning models (LRMs), e.g., DeepSeek-V3.2-Think \citep{deepseekai2025deepseekr1incentivizingreasoningcapability}, to synthesize data with explanations. 
We provide the LLM with the prompt that consists of the given document $doc$ and a claim $c$ from open-source training datasets \citep{lei-etal-2025-factcg}, allowing it to autoregressively provide its chain-of-thought (CoT) $\hat {cot}$, an explanation $\hat e$, and its own predicted label $\hat y$.
By doing so, we can obtain a synthesized sample $\hat{s}$ used for the cold-start SFT stage.
% The used prompts are shown in the Appendix \ref{}.

\subsubsection{Data Filtering}
\label{sec:filtering}
However, even if we apply well-designed prompts, the synthesized data without quality control could still be noisy or useless. 
Thus, we propose a well-defined strategy to avoid low-quality samples without human effort.
We consider three dimensions to ensure data quality, including (1) label correctness, (2) explanation quality, and (3) data diversity.

\noindent
\textbf{Label Correctness.}
For each synthesized sample, we first compare the predicted label $\hat y$ from the LLM with the ground-truth label $y_{\text{gt}}$ provided in the original dataset.
If the two labels are inconsistent, we directly discard the sample $\hat{s}$ along with the generated CoT $\hat{cot}$ and explanation $\hat{e}$. 
Formally,
\begin{align}
    F_{\textit{label}}(\hat{s}) = \mathbb I \Bigl\{ & \hat y=y_{\text{gt}} \Bigr\},
\end{align}
where $\mathbb I$ is the indicator function for filtering low-quality data that do not match the target.
If the label from the LLM is incorrect, the related CoT and explanation may appear coherent, but they are internally aligned with an incorrect prediction.
Including these samples would cause the model to learn incorrect patterns, which would reduce its detection effectiveness and explanation quality.

\noindent
\textbf{Explanation Quality.}
After ensuring the label correctness, we further focus on the explanation quality to prevent low-value or misleading explanations from the training data.
We evaluate the quality of explanations by testing whether they can help the model $\gM$ used for training (e.g., Llama-3.1-8B-Instruct) to make correct predictions.
Specifically, we first measure the model’s perplexity for the ground-truth label using only the document $doc$, claim $c$, and the synthetic CoT $\hat{cot}$: 
\begin{align}
\mathrm{PPL}_{\text{w/o. exp}} 
= \mathrm{PPL}_{\gM}(y_\text{gt} \mid doc, c, \hat{cot}),
\end{align}
which indicates the model’s confidence in generating the correct label.
We then include the synthesized explanation $\hat{e}$ as the input and compute the model’s perplexity again, i.e.,
\begin{align}
\mathrm{PPL}_{\text{w. exp}} 
= \mathrm{PPL}_{\gM}(y_\text{gt} \mid doc, c, \hat{cot}, \hat{e}),
\end{align}
which reflects the model’s confidence in generating the correct label based on the tested explanation.
We retain only the samples $\hat{s}$ with explanations that lower model perplexity on correct labels:
\begin{align}
F_{\textit{exp}}(\hat{s}) 
= \mathbb{I}\{\mathrm{PPL}_{\text{w. exp}} < \mathrm{PPL}_{\text{w/o. exp}}\},
\end{align}
where $\mathbb I$ is the indicator function for filtering data with low-quality explanations.
This indicates that the explanation makes the model more confident in the correct answer, showing that the explanation is both informative and high-quality.
In this way, our method is able to filter out low-quality explanations, ultimately ensuring that the cold-started model can provide high-quality explanations.

\noindent
\textbf{Data Diversity.}
Although filtering for label correctness and explanation quality improves the reliability of individual samples, it may also lead to distribution bias, where the retained data focus on specific tasks and hallucination patterns, ultimately limiting the model’s cross-task generalization.
For instance, the filtering for label correctness can retain too many easy samples, reducing the model’s abilities in complex hallucination scenarios.

Thus, we consider the diversity of the given document $doc$ and claim $c$, since faithfulness hallucinations arise from their semantic relationship.
We adopt a clustering-based approach to preserve data diversity, which can identify semantically close document-claim pairs $(doc, c)$ and form clusters for different types of data.
For each $(doc, c)$ pair from the sample $\hat{s}$, we first use a sentence embedding model to map it to a dense vector.
We utilize the obtained embeddings to employ the $K$-Medoids algorithm \citep{park2009simple} and cosine similarity to get different clusters and their corresponding medoids, i.e., the most centrally located samples in the clusters.
Then, we use the $K$ medoids to construct a probe set $\mathcal{S}_p=\{\hat{s}'_1, ..., \hat{s}'_{K} \}$, then utilize this set to evaluate whether a tested sample $\hat{s}$ can help diverse samples within probe set $\mathcal{S}_p$ towards correct labels.
Specifically, we first infer each probe sample into the model $\gM$ and compute the perplexity of the ground-truth labels:
\begin{align}
\mathrm{PPL}(\hat{s'_i}) = \mathrm{PPL}_{\gM}(\hat{y}'_{i} \mid doc'_{i}, c'_{i}, \hat{cot}'_i,\hat{e}'_i),
\end{align}
where $\hat{s}'_i = (doc'_i, c'_i, \hat{cot}'_i, \hat{e}'_i, \hat{y}'_i)$ denotes the $i$-th sample in probe set $\mathcal{S}_p$.
Next, we incorporate the candidate sample $\hat{s}$ as an in-context demonstration and recompute the perplexity:
\begin{equation}		
		\resizebox{0.89\hsize}{!}{$\begin{aligned}
            \mathrm{PPL}(\hat{s'_i} \mid \hat{s}) = \mathrm{PPL}_{\gM}(\hat{y}'_{i} \mid \hat{s}, doc'_{i}, c'_{i}, \hat{cot}'_i,\hat{e}'_i),
        \end{aligned}$}
\end{equation}
where a decrease in perplexity indicates that $\hat{s}$ provides complementary information that helps the model better predict the correct label for sample $\hat{s}'_i$.
Finally, we count the number of probe samples whose perplexity decreases and retain $\hat{s}$ if it improves a sufficient portion of the probe set:
\begin{equation}	
\label{eq:k-medoids}
		\resizebox{0.89\hsize}{!}{$\begin{aligned}
        F_{\textit{div}}(\hat{s}) = \mathbb{I} \Biggl \{
        \bigl| \{ \hat{s}'_i \in \mathcal{S}_p \mid \mathrm{PPL}(\hat{s}'_i \mid \hat{s}) < \mathrm{PPL}(\hat{s}'_i) \} \bigr|
        \ge \frac{K}{2}
        \Biggr \},
        \end{aligned}$}
\end{equation}
where $\mathbb{I}$ is the indicator function.
In this way, we can ensure that the retained samples have a positive impact across different types of data. 
Consequently, training the model on such diversified and informative samples enhances its ability to maintain strong performance across different tasks.

\noindent
\textbf{Fine-tuning.}
Finally, we apply these three proposed filtering criteria to ensure the data quality, then fine-tune the model on quality-checked training data $\mathcal D$, to get the initialized detection model:
\begin{equation}
		\resizebox{0.87\hsize}{!}{$\begin{aligned}
        \mathcal L_\mathrm{SFT} = - \mathbb E_{\hat{s}\sim\mathcal D}[\log\gM(\hat{cot}, \hat{e}, y_\text{gt} \mid doc, c)].
            \end{aligned}$}
\end{equation}

Thus, the model is equipped with the ability to detect hallucinations and generate explanations.

\subsection{Reinforcement Learning Training}
\label{rl_stage}
The SFT-initialized model can easily memorize the simple training samples and struggles to generalize to complex detection tasks.
Also, the model may generate correct explanations but often lacks clarity or informativeness, as it is trained to imitate training data rather than explicitly optimize for explanation quality.
To further enhance effectiveness and trustworthiness, we frame it as a rule-based RL problem and propose well-designed rewards from prediction correctness and explanation quality.

\subsubsection{Reinforcement Learning Protocol}
\label{gppo}
For the RL training of LLMs, policy optimization methods such as PPO \citep{ppo} and GRPO \citep{shao2024deepseekmathpushinglimitsmathematical} have been well-explored. 
Given the advantages of GRPO, e.g., eliminating the need for a reward model, we utilize the GRPO algorithm to optimize our 
model $\gM_\text{ours}$.

For each document-claim pair $(doc, c)$, the detection model generates a group of $G$ explanations $\{e_1, \dots, e_G\}$, and $G$ candidate corresponding predictions $\{p_1, \dots, p_G\}$. 
Each output is evaluated using a designed composite rule-based reward (\S~\ref{sec:reward}).
GRPO utilizes the relative performance of candidates within the group to compute an advantage $A_i$ for each output, guiding policy updates according to the following objective:
\begin{equation}		
\label{grpo_1}
		\resizebox{0.87\hsize}{!}{$\begin{aligned}
            \mathcal L_\mathrm{GRPO}(\gM_\text{ours}) &= \mathbb{E}_{(doc,c), \{e_i,p_i\} \sim \gM_\text{old}} \left[ \frac{1}{G} \sum_{i=1}^G \mathcal{L}_i - \beta \mathbb{D}_{KL}(\gM_\text{ours} || \gM_\text{ref}) \right],
		\end{aligned}$}
\end{equation}
\begin{equation}		
\label{grpo_2}
		\resizebox{0.75\hsize}{!}{$\begin{aligned}
            \mathcal{L}_i &= \min \left( w_i A_i, \text{clip}(w_i, 1 - \epsilon, 1 + \epsilon) A_i \right),
		\end{aligned}$}
\end{equation}

\noindent
where $w_i = \frac{\gM_\text{ours}(e_i,p_i |doc,c)}{\gM_\text{old}(e_i,p_i |doc,c)}$, $\gM_\text{ref}$ is the reference policy (i.e., the initialized model), $\gM_\text{old}$ is the policy before the update, $\epsilon$ and $\beta$ are hyperparameters for the update step and divergence regularization, and $A_i$ is estimated advantage within the group. 

\subsubsection{Reward Design}
\label{sec:reward}
Having a well-designed reward is key to the effectiveness of RL training \citep{kimik2openagentic}.
An intuitive method is to use a correctness reward to check whether the prediction from the model is correct, ensuring the models can achieve better detection capabilities.
However, this method cannot ensure that the generated explanations are high-quality, as the training signal only explicitly reinforces accurate hallucination detection.
Meanwhile, directly evaluating the quality of free-form explanation via the rule-based verification continues to pose an unresolved challenge \citep{openai2025deep}.
To achieve the balance of trustworthiness and effectiveness, we introduce a prediction correctness reward to improve detection performance and an explanation quality reward to enhance the informativeness and clarity of generated explanations.

\noindent
\textbf{Prediction Correctness Reward.}
This reward assesses whether the detection prediction $y_\text{pred}$ from the model matches the ground-truth answer $y_\text{gt}$, ensuring that the training signal explicitly reinforces accurate hallucination detection. Formally,
\begin{equation}
R_{\text{pred}} = 
\begin{cases} 
1 & \text{if } y_\text{pred} = y_\text{gt}, \\ 
0 & \text{otherwise}.
\end{cases}
\end{equation}

In this way, we can further enhance the model’s prediction accuracy beyond SFT, leading to more reliable detection across diverse scenarios.

\noindent
\textbf{Explanation Quality Reward.}
Directly evaluating the quality of free-form content via the rule-based verification remains challenging.
Thus, we attempt to use the proposed explanation quality reward to evaluate it implicitly.
Specifically, we thoroughly assess a generated explanation by checking if it can help a novice-level model $\gM_\text{nov}$ (e.g., Llama-3.1-8B-Instruct) correctly predict the ground-truth answer. 
The idea behind this reward is that if the generated explanation $e$ enables a novice-level model to generate the correct prediction, it indicates that the explanation is sufficiently coherent and informative for conveying the relevant evidence. Formally,
\begin{equation}
R_{\text{exp}} =
\begin{cases}
1, & \text{if } y_{\text{pred}}^{\,\gM_{\text{nov}}}(doc, c, e) = y_\text{gt}, \\
0, & \text{otherwise},
\end{cases}
\end{equation}
where $y_{\text{pred}}^{\gM_{\text{nov}}}(doc, c, e)$ denotes the final binary prediction produced by the novice-level model conditioned on the provided document $doc$, claim $c$, and generated explanation $e$.
This ensures that only high-quality explanations that are sufficiently coherent and informative are rewarded.

\noindent
\textbf{Format Reward.}
To enforce the desired output format, we assign a format reward to evaluate whether the whole generated response contains the proper tags described in the prompt.
Formally,
\begin{equation}
R_{\text{format}} = 
\begin{cases} 
1, & \text{if correct formatting,} \\ 
0, & \text{if incorrect formatting}.
\end{cases}
\end{equation}

\noindent
\textbf{Final Reward.}
Finally, we use the sum of these three rewards as the final composite reward $R_{\text{final}}$:
\begin{equation}
R_{\text{final}} = R_{\text{pred}} + R_{\text{exp}} + R_{\text{format}}.
\end{equation}

By doing so, we can leverage the well-designed rewards to improve both the detection performance and explanation quality, achieving a distinctive balance of effectiveness and trustworthiness.

%% file: Files/4_Experiment.tex
\section{Experiments}
\label{sec:experiment}
In this section, we conduct experiments and analyses to show the advantages of our FaithLens.

\input{Tabs/main_result}

\noindent
\subsection{Experiment Settings}

\noindent
\textbf{Evaluation.}
We use the cleaned version of LLM-AggreFact \citep{tang-etal-2024-minicheck} and HoVer \citep{jiang-etal-2020-hover} from \citet{seo2025verifying} as benchmarks.
This is because \citet{seo2025verifying} point out that the original version of LLM-AggreFact and HoVer contains 9.1\% ambiguous examples and 6.6\% mislabeled instances.
Also, the used LLM-AggreFact benchmark contains 11 different faithfulness hallucination detection tasks, such as summarization, RAG, and dialogue, to fully evaluate the effectiveness and generalization.
HoVer benchmark further focuses on more complex multi-hop reasoning tasks.
We apply \textit{macro-F1} as our metric, following \citet{seo2025verifying} for a fair comparison.
More details are shown in Appendix \ref{appendix:datasets}.

\noindent
\textbf{Baselines.}
We compare several baselines, including \textbf{(1) The Advanced LLMs}: We evaluate the most advanced LLMs, including GPT-4o, o1, GPT-4.1, o3-mini, o3, GPT-5.2, DeepSeek-V3.2, Llama-3.1-405B-Inst, and Claude-3.7-Sonnet \citep{Claude3S}.
\textbf{(2) Specialized Detection Models:} 
We also compare open-source detection models.
AlignScore \citep{liu2023aligning} trains a 355M-parameter detection model on 4.7M data from 7 different tasks. 
MiniCheck \citep{tang-etal-2024-minicheck} uses Llama-3.1-405B-Inst to synthesize 35K closed-source data and train a 7B model.
% proposes a data synthesis pipeline and uses 35K closed-source data synthesized from Llama-3.1-405B-Inst to train a 7B model.
FactCG \citep{lei-etal-2025-factcg} uses the context graph to generate complex multi-hop synthetic data to train a 435M-parameter model.
ClearCheck \citep{seo2025verifying} uses 57K ANLI examples, 25K closed-source data, and CoTs distilled from Llama-3.1-405B-Inst to train Llama-3.1-8B-Inst with multi-task training.
More details are shown in Appendix \ref{appendix:baseline}.

\noindent
\textbf{Implementation Details.}
For a fair comparison with previous works \citep{seo2025verifying}, our main experiments are conducted on Llama-3.1-8B-Inst.
For training FaithLens, we use the same training data as FactCG \citep{lei-etal-2025-factcg}, as it is based on public data instead of private ones.
Specifically, we utilize the same ANLI \citep{nie-etal-2020-adversarial} subset, C2D, and D2C sets following \citet{lei-etal-2025-factcg} as our initial SFT data, then use our explanation synthesis and filtering strategies, and finally apply SFT on the filtered data.
For the RL stage, we use the CG2C-MHQA and CG2C-Doc sets from \citet{lei-etal-2025-factcg} to train our SFT-initialized model.
In this way, we use the same data as FactCG \citep{lei-etal-2025-factcg} throughout the training process, without introducing additional data.
For computing our explanation quality reward (\S~\ref{sec:reward}), we also use Llama-3.1-8B-Inst as our novice-level model.
We use DeepSeek-V3.2-Think instead of other advanced LRMs (e.g., o3) to synthesize data (\S~\ref{sec:data_syn}), as these models do not allow us to access the CoT content.
More details are shown in Appendix \ref{appendix:imp}, e.g., the sentence embedding model used for data diversity (\S~\ref{sec:filtering}) and hyperparameters.

\subsection{Results}
\label{sec:results}
\noindent
\textbf{Effectiveness Results.} 
As shown in Table \ref{tb:main}, our FaithLens achieves SOTA overall performance on 12 different tasks.
Compared with specialized models, FaithLens not only achieves better results on cross-task scenarios (LLM-AggreFact), but also significantly improves the performance in the complex reasoning detection task (HoVer).
% Compared with 
Meanwhile, FaithLens can achieve better performance than advanced LLMs with much lower cost, e.g., GPT-4.1 and o3.
It shows strong generalization abilities, achieving the lowest standard deviation and the most stable performance across tasks.
% Meanwhile, our training process shows 

\input{Tabs/trustworthiness}
\input{Tabs/infer_efficiency}

\noindent
\textbf{Explainability Results.}
We further evaluate the quality of generated explanations using GPT-4.1 as a judge to show the trustworthiness.
To ensure the correctness and usability of the explanations, we only evaluate the explanations corresponding to the samples that were correctly predicted by the model.
To obtain explanations from advanced LLMs, we adjust the prompts used in effectiveness experiments to require models to generate the explanations before giving their predictions, as FaithLens, which has little to no effect on the model’s prediction performance.
Most of the specialized models (e.g., MiniCheck) treat hallucination detection as a binary classification task and cannot provide explanations.
One exception is ClearCheck, which first generates a CoT and then produces the prediction. 
Thus, we use the CoT from ClearCheck to evaluate its explainability.
Specifically, we consider three dimensions for explanations: readability, helpfulness, and informativeness.
As shown in Table \ref{tb:trustwortheiness}, FaithLens can produce high-quality explanations even compared to advanced LLMs.
This is because our designed data filtering strategy can ensure the quality of explanations used for the SFT stage.
Our explanation quality reward requires LLMs to generate fluent and helpful explanations for a novice model, which further optimizes the quality.
In comparison, unsupervised CoT content from FaithLens cannot serve as high-quality explanations, further showing the effectiveness of our design.
More details can be found in Appendix \ref{appendix:exp}.

\input{Tabs/training_efficiency}

\input{Tabs/aba}
\noindent
\textbf{Efficiency Results.}
As shown in Table \ref{tab:infer_efficiency}, we compare the inference cost with advanced API-based LLMs.
Our proposed FaithLens delivers SOTA performance with the lowest cost, achieving the balance of effectiveness and efficiency.
We also show the comparison of specialized models as shown in Table \ref{tab:training_efficiency}.
FaithLens can achieve reliable performance and provide the corresponding explanations without relying on closed-source and private data.
With our data filtering strategy, our method can efficiently utilize data to achieve better performance. 
In this way, our FaithLens can achieve efficiency in both inference cost and training data.

\subsection{Analysis}
\label{sec:analysis}
\noindent
\textbf{Ablation Study.}
We conduct an ablation study to show the effectiveness of our methods in Table \ref{tab:ablation}.
The results reveal that each of our designed components can significantly enhance the model.
For our data filtering strategy, we find that each considered dimension plays its expected role.
Specifically, the label correctness filtering affects the model’s prediction performance. 
The explanation quality filtering influences the model’s explainability, and the data diversity filtering impacts the consistency of the model’s cross-task performance.
Meanwhile, the proposed rule-based RL stage with a composite reward can further enhance the performance and explainability of the SFT-initialized model.
The designed explanation quality reward effectively improves the quality of corresponding explanations and enhances the final model performance.
More detailed results can be found in Appendix \ref{appendix:abla}.

\input{Tabs/claim}

\begin{figure}[t]
    % \vspace{1.5mm}
    % \hspace{-5mm}
    \centering
    \includegraphics[width=0.3\textheight]
    {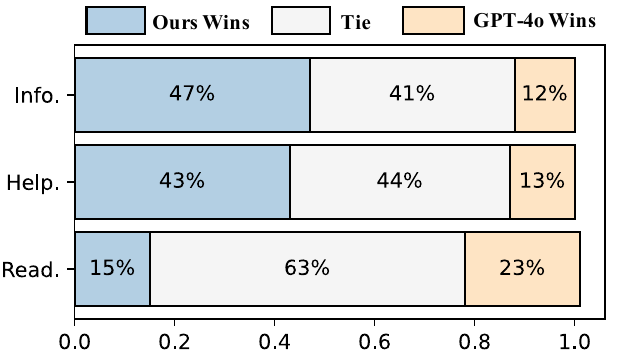}
    \caption{
    \textbf{Human Evaluation.} We compare the explanations from FaithLens and GPT-4o on 120 samples.}
    \label{fig_human_eval}
% \end{wrapfigure}
\end{figure}

\noindent
\textbf{Claim Decontextualization and Claim Decomposition Study.}
We revisit two typical stages in detection pipelines: claim decontextualization and decomposition.
Decontextualization \citep{choi-etal-2021-decontextualization} aims to address coreference and ellipsis in claims, which may make claims difficult to ground.
% Decomposition \citep{min2023factscore} tries to decompose each claim into atomic facts and use the detection model to predict the label for each atomic fact.
Decomposition \citep{min2023factscore} tries to decompose each claim into atomic facts.
If all atomic facts are supported by the document, then the claim is supported; otherwise, the claim is not supported.
We use GPT-4.1 to conduct these two operations, then use the modified claims as new inputs.
As shown in Table \ref{tab:claim}, decontextualization and decomposition are not needed for our model as FaithLens can effectively capture the context-dependent relations.
% Also, we can find that claim decomposition can further improve the performance of FaithLens.
Details can be found in Appendix \ref{appendix:claim}.

\noindent
\textbf{Human Evaluation.}
We conduct the human evaluation for the explanations from FaithLens and GPT-4o on 120 selected samples.
For each comparison, the final result is determined by majority voting for three dimensions: readability, helpfulness, and informativeness. 
Results from Figure \ref{fig_human_eval} show the effectiveness of our method. 
Details are shown in Appendix \ref{appendix:human}, e.g., evaluation principles.

\input{Tabs/backbones}

\noindent
\textbf{Generalization Across Foundation Models.}
As shown in Table \ref{tab:found_models}, using our designed process to train the detection model on different foundational models, e.g., Qwen-2.5-Inst \citep{qwen2.5} and Llama-3.1-Inst, can consistently improve performance compared to the original ones. 
% As shown in Table \ref{tab:found_models}, using our designed training process can consistently improve performance compared to the original ones, including both Qwen-2.5-Inst \citep{qwen2.5} and Llama-3.1-Inst.

\noindent
\textbf{Parameter Study, Variant Methods Testing, and Case Study.}
We also perform these additional analyses in the Appendix \ref{appendix:para}-\ref{appendix:case} to show the effectiveness.

%% file: Tabs/main_result.tex
\begin{table*}
\centering  
\resizebox{1\hsize}{!}{
\begin{tabular}{lcccccccccccccc}
\toprule
% \toprule
 \multirow{2}{*}{\textbf{{Model}}}&

    \multirow{2}{*}{\begin{tabular}[c]{@{}c@{}}\textbf{{Agg-}}\\ \textbf{{CNN}}\end{tabular}} &

    \multirow{2}{*}{\begin{tabular}[c]{@{}c@{}}\textbf{{Agg-}}\\ \textbf{{XSum}}\end{tabular}} &

  \multirow{2}{*}{\begin{tabular}[c]{@{}c@{}}\textbf{{Claim}}\\ \textbf{{Verify}}\end{tabular}} &

  \multirow{2}{*}{\begin{tabular}[c]{@{}c@{}}\textbf{{Expert}}\\ \textbf{{QA}}\end{tabular}} &

  \multirow{2}{*}{\begin{tabular}[c]{@{}c@{}}\textbf{{FC-}}\\ \textbf{{GPT}}\end{tabular}} &

  \multirow{2}{*}{\textbf{{LfQA}}} &

  \multirow{2}{*}{\begin{tabular}[c]{@{}c@{}}\textbf{{RAG}}\\ \textbf{{Truth}}\end{tabular}} &

  \multirow{2}{*}{\textbf{{Reveal}}} &

    \multirow{2}{*}{\begin{tabular}[c]{@{}c@{}}\textbf{{Tofu-}}\\ \textbf{{MediaS}}\end{tabular}} &

    \multirow{2}{*}{\begin{tabular}[c]{@{}c@{}}\textbf{{Tofu-}}\\ \textbf{{MeetB}}\end{tabular}} &

  \multirow{2}{*}{\textbf{{Wice}}} & \multirow{2}{*}{\textbf{{HoVer}}} & \multicolumn{2}{c}{
    \begin{tabular}[c]{@{}c@{}}
        \textbf{Overall}
    \end{tabular}
}  \\
% \cmidrule(lr){14-15}
\cline{14-15}
 &  &  &  &  &  &  &  &  &   &   & & &  \textbf{Std ($\sigma$) $\downarrow$} & \textbf{Avg ($\mu$) $\uparrow$}   \\
\midrule
\rowcolor{mygray} \multicolumn{15}{c}{\cellcolor{myyellow} \textbf{The State-of-the-Art LLMs}} \\
GPT-4o & 62.3	&74.9	&78.3	&68.3&	86.0	&75.0	&81.8	&86.9	&71.5	&76.9	&77.9 &	73.6	&7.0 &76.1 \\
o1  &68.3&	76.7	&77.1&	72.3&	85.0	&76.1	&79.6&	85.9	&65.8&	76.1&	78.7&	79.9	&	5.9&76.8  \\
DeepSeek-V3.2-Non-Think &75.5	&65.5&	75.4&	74.4	&87.9	&72.9	&80.8	&91.0	&65.5	&82.9&	72.7&	76.7 &7.8 &	76.8   \\
DeepSeek-V3.2-Think &86.8	&76.8&	88.0	&80.7&	88.0	&77.4&	85.9	&92.1	&83.5&	91.4&	81.8&	80.0& 	5.1	&84.4 \\
Claude-3.7-Sonnet & 75.6&	73.6&	83.7	&74.4&	86.9&	86.0&	87.0	&88.0	&85.4&	84.0	&86.0&	80.2 & 5.3& 82.6 \\
% Llama-3.1-70B-Inst& 56.3&	62.8	&68.3&	62.7&	81.9	&64.0&	72.9&	81.8&	62.0 &	78.3	&71.4	&79.0 &	8.7	&70.1 \\
Llama-3.1-405B-Inst & 65.5&	71.6&	80.7&	68.8&	82.0&	76.3&	80.7&	84.0 &	67.3&	78.8&	72.5	&81.6  &	6.4 &	75.8 \\
GPT-4.1 &74.1	&73.6&	81.6	&80.3&	91.3&	81.1	&89.1&	93.2	&75.9	&86.3	&86.4	&82.6 &6.5 &	83.0 \\
o3-mini &64.4&	81.5&	80.2&	73.0&	86.0&	81.5	&80.7&	84.8	&77.1&	78.3	&84.0&	78.5& 	5.9&	79.2 \\
o3 &67.8	&77.2&	83.3	&79.6	&86.9&	87.7	&80.6	&92.2&	82.9	&83.8	&82.1&	81.1&	6.0&	82.1 \\
GPT-5.2 & 87.9&	72.7&	88.6&	82.6&	93.5	&89.9	&89.2	&93.3&	85.4	&79.8	&87.8&	82.9&	5.9 &	86.1\\
\midrule
\rowcolor{mygray} \multicolumn{15}{c}{\cellcolor{myyellow} \textbf{Specialized Detection Models}} \\
AlignScore &45.7	&68.0	&79.8&	75.0	&83.7&	86.6	&83.6&	92.2&	75.8&	76.5&	67.3&	73.3& 	12.0 &	75.6\\
FactCG &76.9	&68.1	&76.2&	75.3&	89.0&	86.5&	79.6&	90.0	&79.1&	71.9&	72.2&	73.1& 	7.0 &	78.2\\
MiniCheck &70.0	&72.7&	85.6	&72.9&	86.8&	89.0&	86.9&	91.0&	74.3&	77.8&	\textbf{85.9}	&74.9&	7.5 &	80.7 \\ 
ClearCheck& 72.8&	78.6&	85.4	&72.7&	87.9	&87.0&	83.8	&87.0&	67.8&	75.8&	81.8	&80.3	& 	6.6 &80.1\\
\rowcolor{blue!5} \textbf{FaithLens} &\textbf{84.9}	&\textbf{79.0}	&\textbf{89.4}	&\textbf{79.6}	&\textbf{92.4}	&\textbf{92.1}	&\textbf{86.8}	&\textbf{92.2}	&\textbf{85.1}	&\textbf{87.2}	&85.6	&\textbf{82.9}	&\textbf{4.6}&\textbf{86.4}   \\
\hdashline[2pt/3pt]
% \rowcolor{blue!5}  $\Delta$ Compared to Vanilla. & \textcolor[rgb]{0.7,0,0}{+41.8} & \textcolor[rgb]{0.7,0,0}{+30.4} & \textcolor[rgb]{0.7,0,0}{+71.3} & \textcolor[rgb]{0.7,0,0}{+25.6} & \textcolor[rgb]{0.7,0,0}{+28.9} & \textcolor[rgb]{0.7,0,0}{+28.2} & \textcolor[rgb]{0.7,0,0}{+22.6} & \textcolor[rgb]{0.7,0,0}{+9.8} & \textcolor[rgb]{0.7,0,0}{+6.9} & \textcolor[rgb]{0.7,0,0}{+10.2} & \textcolor[rgb]{0.7,0,0}{+7.3} & \textcolor[rgb]{0.7,0,0}{+6.4} & \textcolor[rgb]{0.7,0,0}{+22.6} & \textcolor[rgb]{0.7,0,0}{+16.2} \\
\rowcolor{blue!5}  $\Delta$ Compared to Llama-3.1-8B-Inst. & \textcolor[rgb]{0.7,0,0}{+41.8}	& \textcolor[rgb]{0.7,0,0}{+30.4}	& \textcolor[rgb]{0.7,0,0}{+25.8}	& \textcolor[rgb]{0.7,0,0}{+29.8}	& \textcolor[rgb]{0.7,0,0}{+22.6}	& \textcolor[rgb]{0.7,0,0}{+46.0}	& \textcolor[rgb]{0.7,0,0}{+34.8}	& \textcolor[rgb]{0.7,0,0}{+14.0}	& \textcolor[rgb]{0.7,0,0}{+34.3}	& \textcolor[rgb]{0.7,0,0}{+24.9}	& \textcolor[rgb]{0.7,0,0}{+37.4}	& \textcolor[rgb]{0.7,0,0}{+19.3}		& \textcolor[rgb]{0,0.7,0}{-6.3}& \textcolor[rgb]{0.7,0,0}{+30.1}\\
\bottomrule
\end{tabular}}
\caption{\textbf{Effectiveness Results.} 
We report experimental results on 12 various datasets from LLM-AggreFact and HoVer benchmarks.
Bold numbers indicate the best performance of specialized detection models.
Our FaithLens simultaneously outperforms other specialized models and advanced LLMs such as GPT-5.2 and o3.}
\label{tb:main}
\end{table*}

%% file: Tabs/trustworthiness.tex
\begin{table}[t]
\scriptsize	
\centering
\resizebox{0.93\linewidth}{!}{
\begin{tabular}{lcccc}
\toprule
\textbf{Model} & \textbf{Read.} & \textbf{Help.} & \textbf{Info.} & \textbf{Avg} \\
\midrule
% \multicolumn{3}{c}{\cellcolor{myyellow} \textbf{Alpaca}}\\
GPT-4o  & 94.4 &	84.8 &	73.0 &	84.1  \\
o1 & 91.8& 	81.6& 	75.4& 	82.9  \\
DeepSeek-V3.2-Non-Think&93.0	&90.6&	84.2 &89.3   \\
DeepSeek-V3.2-Think &94.4&	92.6&	83.0	&90.0  \\
Claude-3.7-Sonnet & 95.7 &	94.6 &	83.7 &	93.5 \\
Llama-3.1-405B-Inst & 90.6	&79.6	&81.0&	83.7 \\
GPT-4.1 &  99.8&	95.2&	83.2&	92.7 \\
o3-mini&  94.6	&88.2&	71.6&	84.8   \\
o3 &97.6	&97.6	&85.2	&93.5  \\
GPT-5.2 &  98.8 &	98.2 & 86.1	&	94.4 \\
\hdashline[2pt/3pt]
ClearCheck &85.2&	79.0&	67.8&	77.3 \\
% \hdashline[2pt/3pt]
CoT from FaithLens &81.4&	76.6&	68.4 & 75.5 \\
\rowcolor{blue!5} \textbf{FaithLens} &\textbf{92.4}	&\textbf{93.4}&	\textbf{85.4}	&\textbf{90.4}  \\
\rowcolor{blue!5}  $\Delta$ Compared to Llama-3.1-8B-Inst. & \textcolor[rgb]{0.7,0,0}{+17.1} &	\textcolor[rgb]{0.7,0,0}{+21.1} &	\textcolor[rgb]{0.7,0,0}{+17.2} &	\textcolor[rgb]{0.7,0,0}{+18.5} \\ 
\bottomrule
\end{tabular}}
\caption{\textbf{Explainability Results.} We use GPT-4.1 to evaluate the generated explanations from three dimensions, including readability (Read.), helpfulness (Help.), and informativeness (Info.).
Bold numbers indicate the best performance of specialized detection models.}
\label{tb:trustwortheiness} 
\end{table}

%% file: Tabs/infer_efficiency.tex
\begin{table}
% \small
\scriptsize	
\centering
\renewcommand{\tabcolsep}{1.2mm}
\begin{tabular}{lc|lc}
\toprule
\begin{tabular}[c]{@{}l@{}}\textbf{Model}\end{tabular} &
  \begin{tabular}[c]{@{}c@{}}\textbf{Cost}\textbf{(\$)}\end{tabular} &
  \begin{tabular}[c]{@{}l@{}}\textbf{Model}\end{tabular} &
  \begin{tabular}[c]{@{}c@{}}\textbf{Cost}\textbf{(\$)}\end{tabular} \\
\midrule
GPT-4o         & 7.3                   &  o1         & 140.6 \\
DeepSeek-V3.2-Non-Think             & 0.8       & o3-mini   & 5.9                \\
DeepSeek-V3.2-Think           & 1.2                   &  o3           & 8.8                  \\
GPT-4.1          & 11.4                        & Claude-3.7-Sonnet              & 14.5                 \\
GPT-5.2 & 15.3 &  \multicolumn{1}{l}{\cellcolor{blue!5}\textbf{FaithLens (8B)}}
& \multicolumn{1}{c}{\cellcolor{blue!5}\textbf{0.1}}\\
% GPT-4-Dectx & 161 & GPT-4-Decmp & 212 \\
\bottomrule
\end{tabular}
\caption{\textbf{Inference Efficiency Results.} 
Inference cost on 1.2K samples from 12 datasets. 
FaithLens delivers SOTA performance with lowest cost (\$ 0.8/GPU-hour).} 
\label{tab:infer_efficiency}
\end{table}

% \rowcolor{blue!5}
% Overall, the most capable models incurs significant cost.

%% file: Tabs/training_efficiency.tex
\begin{table}[t!]
\scriptsize	
\centering
\renewcommand{\tabcolsep}{1.1mm}
\begin{tabular}{lcccc}
\toprule
\textbf{Model}  & \textbf{\# Data} & \textbf{Data Source} & \textbf{Is Explainable?} \\
\midrule
AlignScore   & 4,700K & Open-source & No \\
FactCG &  52K & Open-source & No \\
MiniCheck    & 35K    & Private & No  \\
ClearCheck  & 82K & Private & Partial \\
\hdashline[2pt/3pt]
\rowcolor{blue!5} \textbf{FaithLens}& \textbf{28K} & \textbf{Open-source} & \textbf{Yes} \\
\rowcolor{blue!5} - w/o. Data Filtering  & 52K & \textbf{Open-source} & \textbf{Yes} \\
\bottomrule
\end{tabular}
\caption{\textbf{Data Efficiency Results.} Comparison of specialized detection models on training data sizes, data source, and explainability.}
\label{tab:training_efficiency}
\end{table}

%% file: Tabs/aba.tex
\begin{table}[t]
\centering
% \small
% \footnotesize
\scriptsize
\resizebox{1\linewidth}{!}{
\begin{tabular}{l c c c}
\toprule
{\multirow{2}{*}{\textbf{Method}}} &
\multicolumn{2}{c}{\textbf{Effectiveness}} &
\multicolumn{1}{c}{\textbf{Explainability}}  \\
\cmidrule(lr){2-3}
\cmidrule(lr){4-4}
 & \textbf{Std $\downarrow$} & \textbf{Avg $\uparrow$} & \textbf{Avg  $\uparrow$} \\
\midrule
Llama-3.1-8B-Inst & 10.9 & 56.3 & 71.9  \\
Direct SFT on 52K Data & 6.1 &	79.1  & N/A \\
\rowcolor{blue!5} \textbf{FaithLens} & \textbf{4.6} & \textbf{86.4} &  \textbf{90.4} \\
\hdashline[2pt/3pt]
- w/o. Cold-start SFT Stage &5.7 &	83.4 & 88.1 \\
- w/o. Data Filtering & 6.7 &	81.2 & 82.3  \\
- w/o. Label Correctness Filtering & 5.3 &	83.5 & 86.0 \\
- w/o. Explanation Quality Filtering &4.8 	&85.8 & 83.4 \\
- w/o. Data Diversity Filtering &6.4 	&85.0 & 89.3 \\ 
\hdashline[2pt/3pt]
- w/o. Rule-based RL Stage& 6.0 &	82.6 &83.8  \\
- w/o. Explanation Quality Reward& 5.1& 	85.7& 84.7 \\

\bottomrule
\end{tabular}
}
\caption{
\textbf{Ablation Study.} N/A means the trained model cannot provide the corresponding explanations.
}
\label{tab:ablation}
\end{table}

%% file: Tabs/claim.tex
\begin{table}[t]
\centering
% \small
% \footnotesize
\scriptsize
\resizebox{1\linewidth}{!}{
\begin{tabular}{l c c c c c c}
\toprule
{\multirow{2}{*}{\textbf{Method}}} &
\multicolumn{2}{c}{\textbf{Original}} &
\multicolumn{2}{c}{\textbf{Decontextualization}} &
\multicolumn{2}{c}{\textbf{Decomposition}}  \\
\cmidrule(lr){2-3}
\cmidrule(lr){4-5}
\cmidrule(lr){6-7}
 & \textbf{Std $\downarrow$} & \textbf{Avg $\uparrow$} & \textbf{Std $\downarrow$} & \textbf{Avg  $\uparrow$} & \textbf{Std $\downarrow$} & \textbf{Avg  $\uparrow$} \\
\midrule
GPT-4o &7.0 	&76.1 &
6.9 &	76.1 &
6.6 &	76.6  \\
o1 &5.9 	&76.8 &
6.0 	&76.5 &
5.6 &	77.2  \\
GPT-4.1 &6.5 &	83.0 &
6.5 &	83.0 &
6.2 	&83.3  \\
o3 &6.0 &	82.1 &
6.0 	&82.0 &
5.7 &	82.5  \\
GPT-5.2 &  5.9 & 86.1 & 5.7 & 	86.0  & 6.1 &	85.9  \\
\hdashline[2pt/3pt]
AlignScore &12.0 &	75.6 &
11.9 	&75.4 &
11.5 &	76.1  \\
FactCG &7.0 &	78.2 &
6.8 &	78.0 &
7.1 &	78.6  \\
MiniCheck &7.5 &	80.7 &
7.5 &	80.6 &
7.3 	&80.8  \\
ClearCheck &6.6 &	80.1 &
6.6 	&80.1 &
6.4 &	80.2  \\
\rowcolor{blue!5} \textbf{FaithLens} & \textbf{4.6} & \textbf{86.4} & \textbf{4.6} &	\textbf{86.4} & \textbf{4.4} 	& \textbf{86.6} \\
\bottomrule
\end{tabular}
}
\caption{
\textbf{Claim Decontextualization and Claim Decomposition Study.
}
We use GPT-4.1 to perform these two operations for claims as new inputs.
% Bold numbers indicate the best performance among the baselines.
}
\label{tab:claim}
\end{table}

%% file: Tabs/backbones.tex
\begin{table}[t]
\centering
% \small
% \footnotesize
\scriptsize
\resizebox{0.86\linewidth}{!}{
\begin{tabular}{l c c c}
\toprule
{\multirow{2}{*}{\textbf{Method}}} &
\multicolumn{2}{c}{\textbf{Effectiveness}} &
\multicolumn{1}{c}{\textbf{Explainability}}  \\
\cmidrule(lr){2-3}
\cmidrule(lr){4-4}
 & \textbf{Std $\downarrow$} & \textbf{Avg $\uparrow$} & \textbf{Avg  $\uparrow$} \\
\midrule
Llama-3.1-8B-Inst & 10.9 & 56.3 & 71.9  \\
Llama-3.1-70B-Inst & 8.7 &	70.1& 83.5  \\
Llama-3.1-405B-Inst & 6.4 &	75.8& 83.7 \\
\rowcolor{blue!5} \textbf{FaithLens-8B} & \textbf{4.6} & \textbf{86.4} &  \textbf{90.4} \\
\hdashline[2pt/3pt]
Qwen2.5-3B-Inst &9.1 &	73.3& 79.3  \\
Qwen2.5-7B-Inst &11.3 &	73.9 &81.7  \\
Qwen2.5-32B-Inst &8.6 &	73.1  &84.2 \\
\rowcolor{blue!5} \textbf{FaithLens-3B}& \textbf{4.9} &	\textbf{83.4} & \textbf{88.3}   \\
\rowcolor{blue!5} \textbf{FaithLens-7B} & \textbf{4.2} & 	\textbf{84.9} & \textbf{90.3} \\

\bottomrule
\end{tabular}
}
\caption{
\textbf{Generalization Across Foundation Models.}
The impact of different backbones of the trained models.
}
\label{tab:found_models}
\end{table}

%% file: Files/2_Related_work.tex
\section{Related Work}
\label{appendix:related_work}
\textbf{Hallucinations in LLMs.}
Hallucinations occur when the generated content from LLMs seems believable but does not match factual or contextual knowledge \citep{ji-survey,rawte2023surveyhallucinationlargefoundation,hit-survey}.
% Such hallucinations can undermine the trustworthiness of LLMs in real-world applications \citep{}.
Hallucinations in LLMs can be categorized into factuality hallucinations \citep{min2023factscore,wei2024longformfactualitylargelanguage} and faithfulness hallucinations \citep{huang2025improvingcontextualfaithfulnesslarge, si2025aligninglargelanguagemodels}.
Factuality hallucinations arise when LLMs rely solely on their parametric knowledge and generate statements that contradict real-world facts \citep{wei2024measuringshortformfactualitylarge}.
Faithfulness hallucinations occur when the model's output is inconsistent with or unsupported by the given input, such as a grounding document and retrieved evidence \citep{wan-etal-2023-faithfulness,zhou-etal-2023-context, bi2024contextdpoaligninglanguagemodels, si-etal-2025-gateau}.
LLMs are prone to faithfulness hallucinations across various settings, generating information that cannot be supported by the given context.
For instance, in retrieval-augmented generation, models may generate supplementary information that is not supported by retrieved documents \citep{xu-etal-2024-knowledge-conflicts}.
Even when provided with gold source text, e.g., summarization and simplification tasks, LLMs still produce inconsistent and hallucinated outputs, exhibiting diverse error patterns across domains \citep{gekhman-etal-2023-trueteacher, li-yu-2025-summary}.
In this work, we focus on faithfulness hallucinations, aiming to train an effective and explainable detection model that can assess whether LLM-generated claims remain faithful to the given context.

\noindent
\textbf{Hallucination Detection.}
There are two main approaches to detecting faithfulness hallucinations in LLM-generated outputs.
One relies on advanced LLMs evaluating the LLM-generated outputs, like SelfCheckGPT \citep{manakul-etal-2023-selfcheckgpt}, or further leveraging Chain-of-Thought (CoT) strategies to improve effectiveness \citep{liu-etal-2023-g, dhuliawala-etal-2024-chain, lei2023chainnaturallanguageinference}.
However, these methods are inefficient for real-world applications because they rely on large and advanced models to achieve reliable performance.
Deploying large open-source models requires substantial computing resources, while using advanced API-based models can be very costly.
To reduce cost, the other focuses on training cost-efficient detection models. 
SummaC \citep{laban-etal-2022-summac} adapts natural language inference (NLI) models for document-level faithfulness evaluation.
However, NLI-based approaches struggle with the diverse error patterns and fine-grained faithfulness hallucinations, limiting their robustness across tasks and domains.
Recent studies thus turn to synthetic data generation for training more capable detection models.
AlignScore \citep{zha-etal-2023-alignscore} develops a unified training framework by integrating a large diversity of data sources, resulting in 4.7M training examples from 7 well-established tasks.
MiniCheck \citep{tang-etal-2024-minicheck} synthesizes training data using advanced LLMs and outperforms previous work.
FactCG \citep{lei-etal-2025-factcg} further improves models by enhancing LLM-generated data complexity using knowledge graphs.
ClearCheck \citep{seo2025verifying} uses synthetic data and multi-task training, enabling the model to engage in CoT reasoning before answering.
However, despite these advances in prediction performance, current models still provide only binary labels without accompanying explanations for real-world users, and often exhibit inconsistent performance across tasks.
In this work, we fill these gaps by creating high-quality synthetic data using well-defined data filtering strategies and a carefully crafted rule-based RL stage. 
This enables us to develop FaithLens, a compact detection model that offers a unique combination of trustworthiness, efficiency, and effectiveness.

%% file: Files/5_Conclusion.tex
\section{Conclusion}
\label{subsection:conclusion}
In this paper, we introduce a cost-efficient and effective model, FaithLens, for detecting faithfulness hallucinations while providing corresponding explanations for real-world users. 
We first synthesize training data with explanations and apply a well-defined data filtering strategy to ensure data quality. 
We then fine-tune the model on these well-curated data as a cold start and optimize it with reinforcement learning, using rewards for both prediction correctness and explanation quality. 
In this way, FaithLens can deliver advanced detection effectiveness across 12 different tasks and offer high-quality explanations at a much lower cost.
Overall, our FaithLens achieves a distinctive balance of trustworthiness, efficiency, and effectiveness.

\section*{Acknowledgements}
We would like to thank the anonymous reviewers for their suggestions.
This work is supported by the National Natural Science Foundation of China (No.T2341003), National Natural Science Foundation of China (No. 62236011), and a grant from the Guoqiang Institute, Tsinghua University.

\section*{Limitations}
Although FaithLens demonstrates strong empirical results and is widely applicable, it still has some limitations. 
In this section, we outline these limitations below and explain why they are beyond the scope of this work.
First, we focus exclusively on textual faithfulness hallucination detection and do not address multi-modal settings. 
Extending our FaithLens to multi-modal settings would require fundamentally different grounding signals and explanation formats, which are beyond the scope of this study. 
To ensure the comparability with prior work, we therefore restrict our investigation to the textual domain.
Also, our FaithLens generates its CoT, explanation, and predicted label sequentially. 
Although this design substantially improves trustworthiness and explainability, it introduces additional inference overhead compared to models of similar size that output only predicted labels.
Finally, following standard practice in existing works, FaithLens outputs only binary labels (faithful vs. hallucinated).
While more fine-grained hallucination categories may benefit real-world applications, current datasets lack a unified taxonomy for such distinctions.
We therefore leave fine-grained hallucination detection as future work.
These limitations reflect deliberate choices made to maintain methodological consistency and ensure a fair evaluation, and we view them as promising avenues for extending FaithLens in future research.

%% file: Files/6_Appendix.tex
\appendix
\section*{Appendix}

\noindent This appendix is organized as follows.

\begin{itemize}  
    % \item In Section~\ref{appendix:related_work}, we detail the related work to comprehensively show our motivation.
    \item In Section~\ref{appendix:datasets}, we go into detail about the datasets used in our experiments.
    \item In Section~\ref{appendix:baseline}, we show the details of the baselines during the evaluation.
    \item In Section~\ref{appendix:imp}, we list the details of the implementation, e.g., hyperparameters.
    \item In Section~\ref{appendix:exp}, we further show the details of the explanation evaluation.
    \item In Section~\ref{appendix:abla}, we report the detailed results of the ablation study, e.g., the detailed results.
    \item In Section~\ref{appendix:claim}, we go into details about claim decontextualization and claim decomposition study, e.g., the detailed results.
    \item In Section~\ref{appendix:human}, we show the implementation details of human evaluation.
    \item In Section~\ref{appendix:backbones}, list the details of the generalization test across the foundation models.
    \item In Section~\ref{appendix:para}, we conduct experiments to explore the impact of hyperparameters.
    \item In Section~\ref{appendix:methods_test}, we conduct fine-grained variant method testing to validate the effectiveness of our proposed designs.
    \item In Section~\ref{appendix:case}, we come up with a practical case study to show the effectiveness of FaithLens.

\end{itemize}

\section{Dataset Details}
\label{appendix:datasets}
We introduce 12 various datasets from both LLM-AggreFact and HoVer for our evaluation.
According to \citet{seo2025verifying}, label ambiguity and annotation errors in the original datasets can significantly impact the evaluations.
\citet{seo2025verifying} point out that 9.1\% of the examples are ambiguous, and 6.6\% are mislabeled.
Thus, \citet{seo2025verifying} further constructs a cleaned and well-labeled version of these two benchmarks.
For a fair comparison, we use the cleaned version following \citet{seo2025verifying} to conduct our experiments.
Specifically, the LLM-AggreFact includes 11 different faithfulness hallucination detection tasks, including:

\noindent
\textbf{Agg-CNN \& Agg-XSum.} 
AggreFact \citep{tang-etal-2023-understanding} is an evaluation benchmark for summarization targeting \textbf{CNN(/DM)} \citep{nallapati-etal-2016-abstractive} and \textbf{XSum} \citep{narayan-etal-2018-dont}.
It focuses on the SOTA sets, where documents are from the original CNN and XSum datasets and summaries are generated from SOTA finetuned summarizers, since their analysis suggests that summaries are more challenging to evaluate for hallucination compared to summaries generated by pre-SOTA summarizers.

\noindent
\textbf{ClaimVerify.}
This dataset \citep{liu-etal-2023-evaluating} evaluates the correctness of responses from four generative search engines in answering user queries. 
The dataset contains annotations on whether check-worthy sentences from the engines’ responses can be fully supported by their cited documents.

\noindent
\textbf{ExpertQA.}
It contains responses from 6 different systems to queries curated by experts from 32 fields \citep{malaviya-etal-2024-expertqa}. 
These systems answer queries either in a closed-book fashion, with/without
in-line citations, or based on retrieved document(s).
For each sentence in the response, the sentence is verified against the concatenation of cited or retrieved document(s), if any.

\noindent
\textbf{FC-GPT.}
FactCheck-GPT \citep{Wang2023FactcheckGPTEF} contains factual consistency annotations for LLMs’ responses to search queries. 
In this dataset, each sentence from LLMs’ responses is first decomposed
into atomic facts, and those atomic facts are then decontextualized so that they can stand alone.

\noindent
\textbf{LfQA.}
LFQA \citep{chen2023understanding} contains LLM-generated responses to questions from the ELI5 dataset \citep{fan-etal-2019-eli5}.
LLMs generate responses based on documents retrieved by humans, models, or randomly selected. 
Human annotators then evaluate each sentence in the LLM-generated responses against the corresponding document set, classifying them into supported or not supported.

\noindent
\textbf{RAGTruth.}
It is a hallucination detection corpus in various tasks within the RAG setting \citep{wu2023ragtruth}. 
It comprises naturally generated responses from diverse LLMs using RAG. 
These responses have undergone meticulous manual annotations at both the individual cases and word levels, incorporating evaluations of hallucination intensity.

\noindent
\textbf{Reveal.} REVERL \citep{jacovi-etal-2024-chain}  is a benchmark that evaluates the correctness of reasoning chains generated by LLMs in the context of open-domain question-answering. 
The dataset includes annotations at the sentence level, covering various aspects of response correctness.

\noindent
\textbf{Tofu-MediaS \& Tofu-MeetB.}
These two datasets are collected from TofuEval \citep{tang-etal-2024-tofueval}.
It is a benchmark for dialogue summarization, targeting MediaSum \citep{zhu-etal-2021-mediasum} and MeetingBank \citep{hu-etal-2023-meetingbank}.
It includes topic-focused dialogue summaries generated by 6 LLMs, with sentence-level annotations by linguists.

\noindent
\textbf{Wice.}
WiCE \citep{kamoi-etal-2023-wice} is a textual entailment dataset that consists of naturally occurring claims from Wikipedia and their cited documents.

To evaluate the performance in complex reasoning scenarios, we include the HoVer benchmark.

\noindent
\textbf{HoVer.}
HoVer is an open-domain, many-hop hallucination detection dataset built upon the Wikipedia corpus.
In HoVer, the claims require evidence to be extracted from as many as four English Wikipedia articles and embody reasoning graphs of diverse shapes. 
Most of the 3/4-hop claims are written in multiple sentences, which adds to the complexity of understanding long-range dependency relations such as coreference.

\section{Baseline Details}
\label{appendix:baseline}
In our work, we compare several baselines, including both advanced LLMs and specialized detection models.
In this part, we will detail the version of these models used and the technical details.

\noindent
\textbf{AlignScore.}
It is an entailment-based model that has been trained on 4.7M data from a wide range of tasks such as NLI, QA, fact verification, and summarization.
We use the strongest, largest, and released model trained based on RoBERTa-Large \citep{liu2019roberta} in our experiments\footnote{https://huggingface.co/yzha/AlignScore}.
Meanwhile, we set the prediction threshold as 0.5, then AlignScore outputs a label of either 0 or 1.

\noindent
\textbf{MiniCheck.}
MiniCheck proposes a data synthesis pipeline to automatically get training samples that reflect the complexity of LLM faithfulness hallucination detection.
MiniCheck introduces Claim to Doc (C2D) and Doc to Claim (D2C) generation technologies to generate synthetic documents that require models to be able to check multiple facts in the claim against multiple sentences each, and to generate claims and pair them with portions of these human-written documents, resulting in C2D and D2C sets.
Combined with the ANLI training subset, MiniCheck-FT5 can outperform all systems of comparable size and reach GPT-4 performance.
The authors further train a 7B-level SOTA model MiniCheck-7B\footnote{https://huggingface.co/bespokelabs/Bespoke-MiniCheck-7B} on 35K private data synthesized from Llama-3.1-405B-Inst based on the proposed C2D and D2C technologies.
In this paper, we compare the MiniCheck-7B in experiments.

\noindent
\textbf{FactCG.}
This work investigates the difference between state-of-the-art synthetic generated claims and real LLM-generated claims. 
Then, FactCG proposes a new synthetic data generation approach, CG2C, that leverages the context graph to generate complex multi-hop claims without relying on LLMs to decide data labels, resulting in the CG2C-MHQA and CG2C-Doc sets.
Then the authors use the same ANLI subset, C2D set, D2C set directly from MiniCheck, along with proposed CG2C-MHQA and CG2C-Doc sets (totaling 52K data) to train FactCG-DBT based on DeBERTa-v3-Large \citep{he2020deberta}.
FactCG-DBT\footnote{https://huggingface.co/yaxili96/FactCG-DeBERTa-v3-Large} leverages this generated data to achieve state-of-the-art performance compared with models of similar parameter size and even outperforms GPT-4-o, which is used to construct the CG2C dataset.
We compare FactCG-DBT in our paper, as it is the only released version of FactCG.
Also, we set the prediction threshold as 0.5 for each task.

\noindent
\textbf{ClearCheck.}
\citet{seo2025verifying} found that a small fine-tuned model underperforms larger models by a huge margin, particularly for instances requiring complex reasoning (e.g., HoVer dataset).
Then the authors introduce a simple method to build synthetic multi-hop detection data based on Wikipedia and Llama-3.1-405B-Inst, and experiments show that fine-tuning the model on this data largely improves its performance on examples from the Hover dataset.
ClearCheck again uses Llama-3.1-405B-Inst to generate direct answers for the given documents and claims, and then CoT reasoning traces on 57K ANLI examples and 25.2K private synthetic multi-hop data as training data.
ClearCheck fine-tunes the Llama-3.1-8B-Inst with multi-task training, enabling the model can provide direct answers or engage in CoT reasoning before answering.
We use the released model of ClearCheck-8B\footnote{https://huggingface.co/just1nseo/ClearCheck-8B} to conduct our experiments.
\citet{seo2025verifying} point out that using CoT or providing direct answers does not give different evaluation results; however, CoT makes the verifier output legible to humans so that possible errors can be detected.
Thus, we report the results from ClearCheck with CoT and use the corresponding CoT as the explanation to measure the explainability and trustworthiness.

Here, we list the API versions of the advanced LLMs we used as baselines, including \textit{gpt-4o-2024-08-06} for GPT-4o, \textit{o1-2024-12-17} for o1, \textit{gpt-4.1-2025-04-14
} for GPT-4.1, \textit{o3-2025-04-16
} for o3, \textit{o3-mini-2025-01-31} for o3-mini, \textit{gpt-5.2-2025-12-11} for GPT-5.2, \textit{deepseek-chat} for DeepSeek-V3.2-Non-Think, \textit{deepseek-reasoner} for DeepSeek-V3.2-Think, and \textit{claude-3-7-sonnet-20250219} for Claude-3.7-Sonnet.

\section{Implementation Details}
\label{appendix:imp}
\textbf{Hyperparameters and Devices.}
For the designed data synthesis stage, we use DeepSeek-V3.2-Think to prepare the training data with explanations and set the temperature to 1.0, as it can offer the generated CoTs used for the SFT stage.
For our data filtering stage, we set the number of clusters $K$ as 10, which is used in data diversity filtering according to Eq.(\ref{eq:k-medoids}).
Also, we use the embedding model Llama-Embed-Nemotron-8B \citep{babakhin2025llamaembednemotron8buniversaltextembedding} to get the clusters, which is based on the Llama-3.1-8B model.
Meanwhile, we report the data remaining after applying the filtering strategy shown in Table \ref{tab:data_num}.
Specifically, we sequentially perform Label Correctness Filtering, Explanation Quality Filtering, and Data Diversity Filtering, resulting in the final filtered numbers. 
This is because by first applying Label Correctness Filtering, a large number of useless samples can be removed, eliminating the need to compute the metrics for explanation quality and data diversity on all samples.
Meanwhile, we only apply the filtering strategy during the SFT stage, as the RL data we selected consists of verified, high-quality, and more challenging samples by \citet{lei-etal-2025-factcg}.
For SFT training, we use the Adam optimizer \citep{kingma2017adammethodstochasticoptimization} to train our model, with a $1 \times 10^{-5}$ learning rate with a weight decay of 0.1, and a batch size of 16, steering the training across 3 epochs.
We conduct our SFT stage with DeepSpeed+ZeRO3 and BF16.
For RL training, the learning rate is set to $1 \times 10^{-6}$ for the actor.
We use a group size $G$ of 7, and the rollout temperature is set to 0.6, which is the same as the temperature during the evaluation stage.
Also, for the novice-level model used to compute the explanation quality reward, we set the temperature to 0.6.
The mini-batch size is set to 16, a total of 112 across 7 GPUs for 2 epochs, the KL-divergence loss coefficient $\beta$ is set to 0.001, and $\epsilon$ is set to 0.2. 
The gamma-decay factor $\alpha$ is set to 0.2.
To enforce the desired output format, we assign a format reward on the whole generated response to evaluate whether it contains the proper three types of XML tags, as shown in Figure \ref{fig:prompt_train_inference}.
Our experiments are conducted on NVIDIA A800 SXM4 80G GPUs.

\input{Tabs/data_num}

\noindent
\textbf{Evaluation.}
% During the evaluation for baselines, we infer them twice to report the final results, e.g., AlignScore, or directly use the results from \citet{seo2025verifying}. 
For baselines that were already evaluated under identical benchmark settings in \citet{seo2025verifying}, we directly report the results from that paper in our Table \ref{tb:main} to ensure consistency with the established benchmark.
For baselines not covered in \citet{seo2025verifying}, e.g., FactCG and GPT-5.2, we reproduce the models and infer them twice under the same experimental settings, and report the corresponding results.
Meanwhile, for FaithLens, we also infer our model twice to obtain stable results.

\noindent
\textbf{Prompt Templates.}
We use the same prompt template as shown in Figure \ref{fig:prompt_train_inference} for training and evaluation of FaithLens.
% During the 
For data synthesis, we use the prompt shown in Figure \ref{fig:prompt_data_syn} to query the DeepSeek-V3.2-Think.
For data filtering, we use the prompt shown in Figure \ref{fig:prompt_exp_filtering} to evaluate the explanation quality, and utilize the prompt in Figure \ref{fig:prompt_diversity_filtering} to evaluate whether a tested sample can help diverse samples towards correct labels in the data diversity filtering.
When computing the explanation quality reward, we use the prompt template shown in Figure \ref{fig:prompt_exp_reward} to assess whether a generated explanation can help a novice model correctly predict the correct answer.

\input{Tabs/detailed_performance}
\section{Explainability Results Details}
\label{appendix:exp}
To assess explanation quality, we use GPT-4.1 as an automatic judge. 
All judgments reported in \S~\ref{sec:results} are obtained by querying the GPT-4.1 API (version \textit{gpt-4.1-2025-04-14}) with default parameters and asking it to score each explanation along the three dimensions described in the main paper (readability, helpfulness, and informativeness). 
The exact prompt template we used to query GPT-4.1 for scoring is provided in Figure \ref{fig:llm-as-a-judge}. 
We report the percentage as the final results.
For baselines that by default only produce a binary prediction (i.e., no explanation), including API-based LLMs and Llama-3.1-Inst-series, we modify the prompt used in \citet{seo2025verifying} so that the models are asked to produce both the explanation and the final binary decision. 
The modified prompt is shown in Figure \ref{fig:prompt_to_generate_exp}.
We note that requiring an explanation before the final answer has little to no effect on the numeric prediction outcome, as shown in Table \ref{tb:main} and Table \ref{tb:effectiveness_check} while making the output legible for downstream explainability evaluation.
At the same time, we also investigate whether there is any bias in our use of LLM-as-a-judge. 
Therefore, we conduct additional experiments using a different LLM as the judge. 
Specifically, we used GPT-5-mini API (\textit{gpt-5-mini-2025-08-07}) as the judge.
As shown in Table \ref{tb:detailed_exp}, we can observe that helpfulness and informativeness scores remain stable, while the readability scores fluctuate, especially for GPT-4.1. 
This indicates that when a model is used to evaluate its own outputs, it tends to assign higher scores for readability. 
Regardless of the LLMs used for scoring, our model consistently achieves improvements and maintains advanced performance compared with API-based LLMs.
We also conduct the human evaluation in Appendix \ref{appendix:human} to demonstrate the effectiveness.
By combining both automatic and human evaluation results, we can find that the explanations from FaithLens can even surpass GPT-4o, especially in terms of helpfulness and informativeness.

\section{Ablation Study Details}
\label{appendix:abla}
We further provide the full results corresponding to the ablation study summarized in Table \ref{tab:ablation} of the main paper. 
The complete results are reported in Table \ref{tb:detail_main} and Table \ref{tb:detailed_exp}.
Meanwhile, we find that directly applying SFT on Llama-3.1-8B-Inst with all the data from FactCG \citep{lei-etal-2025-factcg} does not greatly improve performance, indicating that simple scaling does not substantially enhance the capabilities of hallucination detection models.
It demonstrates the necessity of introducing a specifically designed method to train the detection model.

% for particular tasks.
% Therefore, it is necessary to introduce 

\section{Claim Decontextualization and Claim Decomposition Study Details}
\label{appendix:claim}
\textbf{Claim Decontextualization.}
During the faithfulness hallucination detection, phenomena like coreference and ellipsis may make sentences difficult to ground out of context.
Previous methods \citep{choi-etal-2021-decontextualization, tang-etal-2024-minicheck} attempt to address this with an explicit decontextualization step.
We prompt GPT-4.1 API (version \textit{gpt-4.1-2025-04-14}) for decontextualization as shown in Figure \ref{fig:claim_1}, using the previous claims as context to expand the claim following \citet{tang-etal-2024-minicheck}. 
More detailed results can be found in Table \ref{tb:detail_main}.

\noindent
\textbf{Claim Decomposition.}
We also experiment with a setting using claim decomposition. 
In this setting, we decompose each claim $c$ into atomic facts $[af_1, af_2, ...,af_{k} ]$ with the prompt from \citet{kamoi-etal-2023-wice, tang-etal-2024-minicheck} and use the detection model to predict the label for each document-facts pair.
If all atomic facts are supported by the document, then the claim is supported, and unsupported otherwise.
There are typically 2-4 atomic facts per claim across datasets.
We prompt GPT-4.1 API (version \textit{gpt-4.1-2025-04-14}) for decomposition as shown in Figure \ref{fig:claim_2}.
As shown in our experiments, using claim decomposition can improve the final results to a certain extent.
However, this approach increases the inference time and costs by a factor of 2-4 for different datasets, depending on the average number of atomic facts per claim. We believe it should not be used until it provides a significant accuracy benefit.
More detailed results about claim decomposition are shown in Table \ref{tb:detail_main}.

\input{Tabs/detailed_exp}

\section{Human Evaluation Details}
\label{appendix:human}
Evaluating free-form content from LLMs remains challenging.
Thus, we conduct a pairwise human evaluation on the 120 samples from 12 different datasets (10 per dataset) used in our evaluation. 
The order of examples and model outputs was randomized to avoid presentation bias.
We assess these samples across three dimensions: readability, helpfulness, and informativeness.
For each comparison, three options are given (FaithLens Wins, Tie, and GPT-4o Wins), and the majority voting determines the final result. 
The participants follow the principles in Figure \ref{fig:prompt_human_eval} to make the decision. 
We invite three participants pursuing bachelor's or master's degrees to compare the explanations generated by the models. 
Before participants begin to make judgments, we describe the principles of our design in detail and ensure that each participant correctly understands the principles. 
If the final result cannot be determined by majority voting, we will hold a discussion among the participants and vote on the result again. 
We compare two models during the evaluation, including FaithLens as our method and GPT-4o as the advanced model.

\section{Generalization Study Details}
\label{appendix:backbones}
We explore the impact of different model backbones shown in \S~\ref{sec:analysis}.
We also report the detailed results in Table \ref{tb:detail_main} and Table \ref{tb:detailed_exp}.
Meanwhile, to get the generated explanations from the initial models, e.g., Llama-3.1-8B-Inst and Qwen-2.5-7B-Inst, we use the same prompt as detailed in Appendix \ref{appendix:exp}.

\input{Tabs/para_variant_exp}

\input{Tabs/para_variant_performance}

\section{Parameter Study}
\label{appendix:para}
In this section, we further conduct a parameter study to evaluate the effectiveness of our designed modules and gain a better understanding of them.

\noindent
\textbf{Study 1: The Impact of Different Numbers of Clusters in Data Diversity Filtering.}
As the only hyperparameter introduced in our proposed method, we further conduct tests on the hyperparameter $K$ introduced in the data diversity filtering stage.
As shown in Table \ref{tb:para_var_exp} and Table \ref{tb:para_var_per}, our designed data diversity filtering is robust to the hyperparameter $K$.
Meanwhile, increasing the number of clusters may introduce additional computing time used for calculating the perplexity score, and does not significantly improve the performance.

% \noindent
% \textbf{The Impact of Different Thresholds in Data Diversity Filtering.}

\noindent
\textbf{Study 2: The Impact of Different Embedding Models in the Data Diversity Filtering.}
We further explore the impact of different embedding models used in data diversity filtering.
We compare Llama-Embed-Nemotron-8B \citep{babakhin2025llamaembednemotron8buniversaltextembedding}, Linq-Embed-Mistral-7B \citep{LinqAIResearch2024}, and Gemini-Embedding-001\footnote{https://ai.google.dev/gemini-api/docs/embeddings}, which are advanced sentence embedding models according to MTEB leaderboard \citep{muennighoff2023mtebmassivetextembedding}.
We can find that using different advanced embedding models achieve stable results.
Therefore, we recommend using advanced open-source embedding models, e.g., Llama-Embed-Nemotron-8B, for implementation to ensure effectiveness and reduce costs.

\noindent
\textbf{Study 3: The Impact of Different Novice-level Models for Explanation Quality Reward.}
When calculating the explanation quality reward, our motivation is that if the generated explanation enables a novice-level model to produce the correct prediction, it indicates that the explanation is sufficiently coherent and informative for conveying the relevant evidence. 
Therefore, we further investigate how to select the novice-level model, and what would happen if an expert-level model is used to compute the explanation quality reward.
We first explore whether the backbone of the novice-level model needs to be consistent with that of the policy model, that is, whether the policy model and the novice-level model should be homologous models \citep{yu2024language,si2025goalplanjustwish}.
As shown in Table \ref{tb:para_var_exp} and Table \ref{tb:para_var_per}, even though both Llama-3.1-8B-Inst and Qwen-2.5-7B-Inst perform poorly on faithfulness hallucination detection and can be considered novice-level models, we can find that using Llama-3.1-8B-Inst as the novice-level model can achieve better performance than using a heterologous model, i.e., Qwen-2.5-7B-Inst.
This may be due to different pre-training data, language styles, or sensitivity to instruction formats, which can result in a particular model being unable to correctly predict labels based on the provided explanation.
Also, by using the homologous novice-level model, we also avoid having the measurement of explanation quality biased by factors other than the novice-level model’s problem-solving skills, such as context windows.
We further test using an expert-level model instead of a novice model for reward computing; specifically, we use DeepSeek-V3.2-Think to perform the experiment. 
We find that using the expert-level model results in suboptimal performance, particularly regarding the quality of the generated explanations. 
This may be because the expert model can ignore the incorrect explanations provided and still predict the correct label.
As a result, low-quality explanations are assigned high rewards, which in turn weakens the quality of the explanations generated by the policy model.

\noindent
\textbf{Study 4: The Impact of Few-shot Prompting for LLM Baselines.}
We additionally conducted few-shot experiments following the demonstration examples provided in \citet{seo2025verifying}. 
As expected, few-shot prompting improves the prediction accuracy of API-based LLMs. 
However, even under the few-shot setting, FaithLens still consistently outperforms strong API models such as GPT-4o and o1 across the evaluated datasets. 
Although the performance gap is partially reduced, the advantage of FaithLens remains substantial. Moreover, as shown in Table \ref{tab:infer_efficiency}, FaithLens requires only approximately 0.1\% of the inference cost of o1. 
This highlights that our advantage is not limited to predictive performance. 
Compared to LLM-based approaches that rely on large proprietary models at inference time, FaithLens achieves competitive or superior performance while maintaining significantly lower computational cost.

\section{Variant Methods Testing}
\label{appendix:methods_test}
In this section, we further explore the variant methods in our designs to gain a better understanding of our proposed FaithLens and design choices, including both SFT and rule-based RL stages.

\noindent
\textbf{Question 1: Can We Provide Explanations Directly Instead of Presenting CoTs First?}
In this work, we require FaithLens to first generate a CoT, then the corresponding explanation, and finally the predicted answer. 
In this paradigm, whether it is possible to directly provide the corresponding explanation without first generating a CoT remains an open question. 
This is because generating a CoT increases inference time, which is critical for real-time hallucination detection.
Therefore, in the SFT stage, we remove the CoT module and retain only the explanation and answer parts of the synthetic data (as shown in Figure \ref{fig:only_exp}) to train the detection model, allowing the model to directly generate the corresponding explanation and the final prediction.
As shown in Table \ref{tb:para_var_exp} and Table \ref{tb:para_var_per}, removing the CoT during the cold start stage greatly impairs the model’s performance and results in a poorly initialized model. 
This indicates that explanations alone cannot serve the same role as the CoT in improving faithfulness hallucination detection performance, and further demonstrates that the process of first generating the CoT and then the corresponding explanation is a well-justified design choice.
We also provide experimental results of directly using the CoTs for SFT as shown in Table \ref {tb:para_var_exp} and Table \ref{tb:para_var_per}. 
The used prompt is shown in Figure \ref{fig:only_cot}.
The experiments show that its effectiveness performance is inferior to using both CoTs and explanations for SFT together. 
This is because our data filtering strategy ensures the quality of explanations, which allows these explanations to help the model achieve better performance.
At the same time, using only CoTs for SFT results in the absence of the explanation generation process, and the content from CoTs is difficult to serve as high-quality explanations.

\noindent
\textbf{Question 2: Why Don't We Use  Perplexity as a Metric for Calculating Rewards During the RL Stage?}
During the data selection stage, especially in explanation quality filtering and data diversity filtering, we design these data filtering methods based on calculating perplexity scores. 
One of the main motivations is that, compared to methods that require the model to perform generation and inference for data filtering, perplexity-based approaches can significantly reduce the cost of the filtering process.
However, during the RL stage and computing the explanation quality reward, we thoroughly assess a generated explanation by checking if it can help a novice-level model correctly predict the ground-truth answer, rather than merely judging whether the novice model succeeds in reducing the corresponding perplexity on the final predicted answers.
Therefore, we further investigate whether this variant approach is effective. 
Specifically, when calculating the explanation quality reward, if the generated explanation successfully reduces the novice-level model's perplexity on the ground-truth answer, a reward of 1 is assigned; otherwise, the reward is 0.
As shown in Table \ref{tb:para_var_exp} and Table \ref{tb:para_var_per}, this variant perplexity-based approach is not as effective as the method we proposed in the main text, i.e., using correctness as a metric for explanation quality reward. 
This may be because reducing perplexity is a simpler task compared to using correctness, which limits the model’s ability to explore more effective policies during the rule-based RL stage and thus prevents it from achieving better performance.

\section{Case Study}
\label{appendix:case}
We conduct the case study for generated explanations from correctly predicted samples in Figure \ref{fig:case_aggre} and Figure \ref{fig:case_hover} to visually show the advantages of FaithLens compared with the advanced LLMs, including GPT-4o and o1.

\noindent
\textbf{Case Study from LLM-AggreFact.}
Figure \ref{fig:case_aggre} illustrates a hallucination detection task involving a compound claim regarding the Federal Lanham Act and the FTC Act. 
FaithLens first summarizes the claim and supporting documents to enhance readability, enabling users to clearly understand the input content. 
Subsequently, rather than simply pointing out the hallucinatory part (the Lanham Act), FaithLens compares it against other pieces of evidence detailed in the documents (such as the Truth in Lending Act, the Fair Credit Reporting Act, etc.). 
By explicitly pointing out that the documents list several specific statutes without mentioning the Lanham Act, FaithLens provides a rigorous, evidence-based explanation for its conclusion, using the omission as strong proof and thereby significantly improving helpfulness and informativeness.
For baselines, GPT-4o provides only a general summary of the document, while o1 directly offers a brief negative judgment.
GPT-4o does not explain the content of the claim, and merely describes what the document includes and notes the absence of mention of the Lanham Act, which severely undermines the readability of the explanation. 
Furthermore, the generated explanation from GPT-4o does not cite evidence from the original text, which reduces its helpfulness and informativeness.
% The generated explanation from o1
The explanation from o1 merely states that the document does not mention the Lanham Act at all, lacking any breakdown of the claim itself and supplementary details from the document.

\noindent
\textbf{Case Study from HoVer.}
As shown in Figure \ref{fig:case_hover}, the explanation from FaithLens demonstrates superior quality compared to the ones from GPT-4o and o1. 
First, the explanation generated by FaithLens demonstrates better readability by restating the claim at the beginning of the explanation, which enhances clarity and allows readers to directly understand the topic under discussion.
It then cites evidence from the document to provide the final conclusion, enhancing the clarity.
Additionally, FaithLens analyzes other atomic facts within the claim and clearly points out that, although there is a hallucination regarding 1940 in the claim, some parts of the claim are correct. 
This prevents users from misunderstanding and increases the informativeness and helpfulness of the generated explanation.
In this scenario, GPT-4o adopts a misleading flow, beginning its explanation by validating the correct definition of animation; this creates initial ambiguity regarding the claim's overall truthfulness. 
The o1 model provides a correct verdict but lacks sufficient explanatory detail to be actionable.

\newpage

\input{Tabs/performance_check}
\input{Tabs/prompt_training}

%% file: Tabs/data_num.tex
\begin{table}[t!]
\scriptsize	
\centering
\renewcommand{\tabcolsep}{1.1mm}
\begin{tabular}{lc}
\toprule
\textbf{Model}  & \textbf{\# Data} \\
\midrule
Initial Whole Data (i.e., - w/o. Data Filtering) & 52,268 \\
Initial SFT Data (i.e., - w/o. Data Filtering) & 35,554 \\
Initial Data For RL & 16,714 \\
\hdashline[2pt/3pt]
Filtered Data from whole Data Filtering & 23,625 \\
Filtered Data from Label Correctness Filtering  &  14,258 \\
Filtered Data from Explanation Quality Filtering & 4,363 \\
Filtered Data from Data Diversity Filtering & 5,004 \\
\hdashline[2pt/3pt]
Final Data For SFT & 11,929 \\
Final Data For RL & 16,714 \\
\rowcolor{blue!5} \textbf{FaithLens}& \textbf{28,643} \\
\bottomrule
\end{tabular}
\caption{\textbf{The Number of The Used Data.} We list the number of filtered data by our proposed filtering strategy and the used data for training FaithLens.}
\label{tab:data_num}
\end{table}

%% file: Tabs/detailed_performance.tex
\begin{table*}[!t]
\centering  
\resizebox{1\hsize}{!}{
\begin{tabular}{lcccccccccccccc}
\toprule
% \toprule
 \multirow{2}{*}{\textbf{{Model}}}&

    \multirow{2}{*}{\begin{tabular}[c]{@{}c@{}}\textbf{{Agg-}}\\ \textbf{{CNN}}\end{tabular}} &

    \multirow{2}{*}{\begin{tabular}[c]{@{}c@{}}\textbf{{Agg-}}\\ \textbf{{XSum}}\end{tabular}} &

  \multirow{2}{*}{\begin{tabular}[c]{@{}c@{}}\textbf{{Claim}}\\ \textbf{{Verify}}\end{tabular}} &

  \multirow{2}{*}{\begin{tabular}[c]{@{}c@{}}\textbf{{Expert}}\\ \textbf{{QA}}\end{tabular}} &

  \multirow{2}{*}{\begin{tabular}[c]{@{}c@{}}\textbf{{FC-}}\\ \textbf{{GPT}}\end{tabular}} &

  \multirow{2}{*}{\textbf{{LfQA}}} &

  \multirow{2}{*}{\begin{tabular}[c]{@{}c@{}}\textbf{{RAG}}\\ \textbf{{Truth}}\end{tabular}} &

  \multirow{2}{*}{\textbf{{Reveal}}} &

    \multirow{2}{*}{\begin{tabular}[c]{@{}c@{}}\textbf{{Tofu-}}\\ \textbf{{MediaS}}\end{tabular}} &

    \multirow{2}{*}{\begin{tabular}[c]{@{}c@{}}\textbf{{Tofu-}}\\ \textbf{{MeetB}}\end{tabular}} &

  \multirow{2}{*}{\textbf{{Wice}}} & \multirow{2}{*}{\textbf{{HoVer}}} & \multicolumn{2}{c}{
    \begin{tabular}[c]{@{}c@{}}
        \textbf{Overall}
    \end{tabular}
}  \\
% \cmidrule(lr){14-15}
\cline{14-15}
 &  &  &  &  &  &  &  &  &   &   & & &  \textbf{Std ($\sigma$) $\downarrow$} & \textbf{Avg ($\mu$) $\uparrow$}   \\
\midrule
\rowcolor{mypink} \multicolumn{15}{c}{\cellcolor{mypink} \textbf{Ablation Study}} \\
Llama-3.1-8B-Inst & 43.1 & 48.6 & 63.6 & 49.8 & 69.8 & 46.1 & 52 & 78.2 & 50.8 & 62.3 & 48.2 & 63.6 & 10.9 & 56.3 \\
Direct SFT on 52K Data  & 72.2 & 70.1 & 77.6 & 76.8 & 85.9 & 87.2 & 80.4 & 90.9 & 78.5 & 75.3 & 78.1 & 76.2 & 6.1 & 79.1 \\
\rowcolor{blue!5} \textbf{FaithLens} &{84.9}	&{79.0}	&{89.4}	&{79.6}	&{92.4}	&{92.1}	&{86.8}	&{92.2}	&{85.1}	&{87.2}	&85.6	&{82.9}	&{4.6}&{86.4}   \\
- w/o. Cold-start SFT Stage &82.9 & 73.9 & 85.8 & 78.3 & 89.1 & 88.8 & 85.3 & 93.3 & 75.3 & 84.7 & 83.3 & 80.3 & 5.7 & 83.4 \\
- w/o. Data Filtering  & 78.2 & 72.5 & 86.9 & 73.5 & 87.9 & 87.7 & 82.7 & 90.9 & 70.9 & 79.2 & 85.3 & 78.6 & 6.7 & 81.2 \\
- w/o. Label Correctness Filtering & 77.7 & 82.4 & 86.0 & 79.0 & 92.4 & 84.1 & 84.5 & 93.2 & 75.9 & 81.0 & 84.4 & 81.2 & 5.3 & 83.5 \\
- w/o. Explanation Quality Filtering & 84.9 & 83.4 & 84.7 & 76.8 & 92.4 & 87.6 & 85.5 & 94.4 & 82.4 & 88.4 & 87.8 & 81.6 & 4.8 & 85.8 \\
- w/o. Data Diversity Filtering & 82.1 & 74.7 & 87.1 & 79.4 & 94.6 & 92.1 & 91.3 & 91.1 & 76.4 & 85.5 & 84.4 & 81.8 & 6.4 & 85.0 \\ 
- w/o. Rule-bsaed RL Stage & 78.1 & 78.0 & 87.2 & 79.2 & 87.0 & 89.7 & 82.9 & 94.4 & 74.1 & 79.8 & 84.4 & 76.9 & 6.0 & 82.6 \\
- w/o. Explanation Quality Reward & 84.9 & 83.1 & 87.1 & 76.3 & 89.1 & 93.2 & 90.2 & 92.1 & 80.4 & 82.1 & 87.8 & 82.1 & 5.1 & 85.7 \\
\midrule
\rowcolor{mypink} \multicolumn{15}{c}{\cellcolor{mypink} \textbf{Claim Decontextualization Study}} \\
GPT-4o & 62.3 & 74.9 & 78.3 & 68.6 & 85.9 & 75.0 & 82.0 & 86.7 & 71.6 & 76.7 & 77.6 & 73.8 & 6.9 & 76.1 \\
o1 & 68.1 &	76.6 	&77.1& 	71.8 &	85.0 &	76.2 	&78.8 	&86.3 	&65.8 &	76.3 &	76.6 	&79.9 &	6.0 &	76.5  \\
GPT-4.1 & 74.1 &	73.6 &	81.6 &	80.3 &	91.3 &	81.1 &	89.1 &	93.2 &	75.9 &	86.3 &	86.4 &	82.6 &	6.5 &	83.0 \\
o3 &  67.6& 	77.3 &	83.2 &	79.2 &	86.9 	&87.6& 	80.6 	&92.1 &	83.2 &	82.7 &	82.1 &	81.3& 	6.0 	&82.0 \\
GPT-5.2 & 87.9&	72.7	&87.6&	81.3	&92.6	&90.3&	91.0 &	90.8&	86.2&	80.2&	88.2&	83.1&	5.7 &	86.0 \\
\hdashline[2pt/3pt]
AlignScore & 45.3 	&67.2 &	79.8 	&75.6 &	83.7 	&86.7 &	82.5 	&91.1 &	75.9 &	77.2 &	67.1 &	72.4 &	11.9 &	75.4   \\
FactCG & 76.4 	&68.6 &	76.2 &	75.7 &	89.0 &	85.6 &	78.3 	&89.7 &	79.0 &	71.4 &	72.5 	&73.2 	&6.8 	&78.0   \\
MiniCheck & 70.0 &	72.7 &	85.6 &	72.9 &	86.8 &	88.9 &	86.9 &	91.0 &	74.3 &	77.8 &	85.6 	&74.9& 	7.5 &	80.6 \\
ClearCheck & 72.8 &	78.6 &	85.4 &	72.7 &	87.9 &	87.2 &	83.8 	&86.7 &	67.8 &	75.8& 	81.8 &	80.3 &	6.6 &	80.1  \\
\rowcolor{blue!5} \textbf{FaithLens} & {84.9}	& {79.0} &	{89.4} 	& {79.6} &	{92.4} &	{92.1} &	{86.8} &	{92.2} &	{85.1} 	& {87.2} &	{85.6} 	& {82.9} & 	{4.6} &	{86.4}  \\
\midrule
\rowcolor{mypink} \multicolumn{15}{c}{\cellcolor{mypink} \textbf{Claim Decomposition Study}} \\
GPT-4o & 63.6 &	75.8 &	78.2 	&68.6 &	85.9 &	75.1 &	82.0 &	86.7 &	72.6 &	77.8 &	78.3 &	74.1 &	6.6 	&76.6  \\
o1 & 69.2 	&78.3 &	77.2 &	72.3 &	85.0 &	76.3 	&79.6 &	85.6 	&66.8 &	76.9 &	79.0 &	80.2 &	5.6 &	77.2  \\
GPT-4.1 & 75.3 &	74.2 &	82.0& 	81.0 &	91.3 	&81.5 &	89.1 	&93.3 	&75.9 &	86.2 &	86.6 	&82.8 &	6.2 &	83.3  \\
o3 &  68.9 	&79.1 &	82.2 &	79.6& 	86.9 &	87.7 	&81.7 &	92.2 &	82.9 &	84.6 	&82.9 	&81.4& 	5.7 &	82.5  \\
GPT-5.2 &87.6	&71.8	&88.6	&82.4	&93.2&	90.1&	89.2&	92.8&	85.1	&80.1&	88.2&	81.6&	6.1 &	85.9 \\
\hdashline[2pt/3pt]
AlignScore &  47.2& 	69.2& 	80.1 &	75.9 &	83.7& 	86.7 	&84.3 	&92.2 &	75.4 &	76.7 &	68.1 &	74.2 	&11.5 &	76.1  \\
FactCG & 78.2 &	69.2 &	76.3 &	76.1 &	89.0 &	87.3& 	78.3 &	91.1 &	79.6 &	71.9 &	72.5 &	73.6 &	7.1 &	78.6  \\
MiniCheck & 72.0 &	74.3 &	86.2 &	71.6 &	86.8 &	88.8 &	87.1 &	90.9 &	74.6 &	77.0 &	85.6 &	74.5 &	7.3 &	80.8  \\
ClearCheck & 73.5 &	78.8 &	86.2& 	73.3 &	87.9 &	87.3 &	83.8 &	85.6 	&68.3 &	75.3 &	82.0 &	80.4& 	6.4 &	80.2  \\
\rowcolor{blue!5} \textbf{FaithLens}& {85.3} 	&{79.5} 	& {89.6} &	{80.0} 	& {92.4} 	& {92.1} &	{86.9} &	{92.0} 	& {85.2} &	{87.4} &	{85.6} 	& {82.9} &{4.4} &	{86.6}  \\
\midrule
\rowcolor{mypink} \multicolumn{15}{c}{\cellcolor{mypink} \textbf{Generalization Study}} \\
Llama-3.1-8B-Inst & 43.1&	48.6	&63.6&	49.8&	69.8	&46.1	&52.0	&78.2&	50.8&	62.3&	48.2&	63.6&	10.9 &	56.3  \\
Llama-3.1-70B-Inst & 56.3&	62.8&	68.3&	62.7&	81.9	&64.0	&72.9	&81.8	&62.0&	78.3&	71.4&	79.0&	8.7 	&70.1  \\
Llama-3.1-405B-Inst &65.5	&71.6&	80.7&	68.8&	82.0&	76.3&	80.7&	84.0&	67.3&	78.8&	72.5&	81.6&	6.4 	&75.8  \\
\rowcolor{blue!5} \textbf{FaithLens-8B}& {84.9}&	{79.0}&	{89.4}&	{79.6}&{92.4}	&{92.1}&{86.8}&	{92.2}&	{85.1}&	{87.2}&	{85.6}	&{82.9}&	{4.6} &	{86.4} \\
\hdashline[2pt/3pt]
Qwen-2.5-3B-Inst & 55.2	&67.9&	78.0&	80.1&	78.1&	78.3&	67.0 &	90.0 	&65.3 &	79.5 &	70.0&	70.6	&9.1 &	73.3  \\
Qwen-2.5-7B-Inst & 52.0&	63.6&	78.4	&80.7	&85.8&	79.5&	68.0 &	92.1 &	62.3& 	70.1 &	82.1&	72.0&	11.3 &	73.9  \\
Qwen-2.5-32B-Inst & 53.8&	65.0&	76.8&	66.1&	83.0&	71.5	&75.0&	82.9&	69.6&	77.8&	81.9&	74.1&	8.6 	&73.1  \\
\rowcolor{blue!5} \textbf{FaithLens-3B}& {82.3}	&{78.3}&	{91.1}&	{77.6}&	{90.2}	&{80.6}&	{85.3}&	{91.1}&	{80.2}&	{83.2}&	{80.3}&	{80.6}&	{4.9} 	&{83.4}  \\
\rowcolor{blue!5} \textbf{FaithLens-7B}&  {83.6} &	 {79.3} &	 {87.6} &	 {78.3} &	 {91.3} &	 {82.1} &	 {87.1} &	 {92.1} &	 {84.2} &	 {85.2} &	 {84.3} &	 {83.1} &	 {4.2} &	{84.9}  \\
\bottomrule
\end{tabular}}
\caption{\textbf{Detailed Effectiveness Results for Ablation Study, Claim Study, and Generalization Study.} 
We report experimental results on 12 various datasets from LLM-AggreFact and HoVer benchmarks.}
\label{tb:detail_main}
\end{table*}

%% file: Tabs/detailed_exp.tex
\begin{table}[t]
\scriptsize	
\centering
\resizebox{0.93\linewidth}{!}{
\begin{tabular}{lcccc}
\toprule
\textbf{Model} & \textbf{Read.} & \textbf{Help.} & \textbf{Info.} & \textbf{Avg} \\
\midrule
\multicolumn{5}{c}{\cellcolor{mypink} \textbf{Using GPT-5-mini as a Judge}} \\
GPT-4o  & 90.2 &	84.6 &	72.5 &	82.4  \\
o1 &  89.2 &	80.9 &	75.6 &	81.9  \\
DeepSeek-V3.2-Non-Think&  90.6 &	90.5 &	85.1 	&88.7  \\
DeepSeek-V3.2-Think & 91.2 	&93.1 &	84.1& 	89.5  \\
Claude-3.7-Sonnet & 90.7 &	93.8 &	83.5 &	89.3  \\
Llama-3.1-405B-Inst & 88.6 &	78.3 &81.6& 	82.8  \\
GPT-4.1 & 95.3 &94.6 &	82.8 &	90.9 \\
o3-mini &   91.8 &	88.3 	&70.8 &	83.6  \\
o3 & 93.2 &	96.9 &	85.5 &	91.9 \\
GPT-5.2 &94.4 & 95.3 &86.3 &92.0 \\
\hdashline[2pt/3pt]
ClearCheck & 83.2 &	78.6 &	68.2 &	76.7  \\
CoT from FaithLens & 79.2 	&75.3 &	67.6 &	74.0  \\
\rowcolor{blue!5} \textbf{FaithLens} & 91.9 &	93.7 &	85.6& 	90.4  \\
\rowcolor{blue!5}  $\Delta$ Compared to Llama-3.1-8B-Inst. &	\textcolor[rgb]{0.7,0,0}{+18.7 } &	\textcolor[rgb]{0.7,0,0}{+23.5 } &	\textcolor[rgb]{0.7,0,0}{+18.0 }  & \textcolor[rgb]{0.7,0,0}{+20.1 }\\ 
% \bottomrule
\midrule
\rowcolor{mypink} \multicolumn{5}{c}{\cellcolor{mypink} \textbf{Ablation Study}} \\
Llama-3.1-8B-Inst & 75.3 &	72.3 &	68.2 &	71.9 \\
Direct SFT on 52K Data  & N/A & N/A& N/A &N/A \\
\rowcolor{blue!5} \textbf{FaithLens} & {92.4} &	{93.4} &	{85.4} 	& {90.4}  \\
- w/o. Cold-start SFT Stage & 90.2 &	90.2 &	83.8 &	88.1 \\
- w/o. Data Filtering  & 85.2 &	82.8 &	78.9 &	82.3  \\
- w/o. Label Correctness Filtering & 88.3 	&88.2& 	81.6 &	86.0  \\
- w/o. Explanation Quality Filtering & 85.5 &	84.3 &	80.3 	&83.4  \\
- w/o. Data Diversity Filtering &91.7 &	91.7 	&84.5& 	89.3  \\ 
- w/o. Rule-bsaed RL Stage & 87.8 &	84.2 &	79.5 &	83.8 \\
- w/o. Explanation Quality Reward & 88.5 &	85.6 &	80.1 &	84.7 \\
\midrule
\rowcolor{mypink} \multicolumn{5}{c}{\cellcolor{mypink} \textbf{Generalization Study}} \\
Llama-3.1-8B-Inst &  75.3 &	72.3 &	68.2& 	71.9  \\
Llama-3.1-70B-Inst & 90.8 &	78.6 &	81.2 &	83.5  \\
Llama-3.1-405B-Inst & 90.6 &	79.6 &	81.0 	&83.7 \\
\rowcolor{blue!5} \textbf{FaithLens-8B}& {92.4} &	{93.4} &	{85.4} &	{90.4}  \\
\hdashline[2pt/3pt]
Qwen-2.5-3B-Inst & 86.7	&78.6&	72.6	&79.3  \\
Qwen-2.5-7B-Inst & 88.6	&80.2&	76.4&	81.7  \\
Qwen-2.5-32B-Inst &  90.4	&81.6	&80.6&	84.2  \\
\rowcolor{blue!5} \textbf{FaithLens-3B}& {90.6}&	{91.6}&	{82.6}	&{88.3}  \\
\rowcolor{blue!5} \textbf{FaithLens-7B}&  {93.5}&	{92.6}	&{84.8}	&{90.3} \\
\bottomrule
\end{tabular}}
\caption{\textbf{Detailed Explainability Results for Using GPT-5-mini as a Judge, Ablation Study and Generalization Study.} We evaluate the generated explanations from three dimensions, including readability (Read.), helpfulness (Help.), and informativeness (Info.).
We use GPT-4.1 as a judge for Ablation Study and Generalization Study.
N/A means that the trained model cannot provide the corresponding explanations.
}
\label{tb:detailed_exp} 
\end{table}

%% file: Tabs/para_variant_exp.tex
\begin{table}[t]
\scriptsize	
\centering
\resizebox{1\linewidth}{!}{
\begin{tabular}{lcccc}
\toprule
\textbf{Model} & \textbf{Read.} & \textbf{Help.} & \textbf{Info.} & \textbf{Avg} \\
\midrule
\rowcolor{mypink} \multicolumn{5}{c}{\cellcolor{mypink} \textbf{Study 1}} \\
\rowcolor{blue!5} \textbf{FaithLens} ($K$=10) & 92.4 	&93.4& 	85.4 &	90.4  \\
- w. Setting $K$ as 6 &93.1	&91.6&	84.6&	89.8   \\
- w. Setting $K$ as 14 & 91.5	&92.3	&84.4&	89.4   \\
- w. Setting $K$ as 20 & 92.2	&91.7&	85.1	&89.7  \\
\midrule
\rowcolor{mypink} \multicolumn{5}{c}{\cellcolor{mypink} \textbf{Study 2}} \\
\rowcolor{blue!5} \textbf{FaithLens} (- w. Llama-Embed-Nemotron-8B)&  92.4 &	93.4 &	85.4 &	90.4 \\
- w. Linq-Embed-Mistral-7B &92.1&	92.8&	84.2	&89.7  \\
- w. Gemini-Embedding-001 &92.2	&92.6&	85.1&	90.0 \\
\midrule
\rowcolor{mypink} \multicolumn{5}{c}{\cellcolor{mypink} \textbf{Study 3}} \\
\rowcolor{blue!5}  \textbf{FaithLens} (- w. Using Llama-3.1-8B-Inst) & 92.4 & 	93.4 & 	85.4 	& 90.4  \\
- w. Using Qwen-2.5-7B-Inst & 90.6&	90.7&	83.9	&88.4  \\
- w. Using DeepSeek-V3.2-Think& 89.2	&88.3	&82.1	&86.5 \\
\midrule
\rowcolor{mypink} \multicolumn{5}{c}{\cellcolor{mypink} \textbf{Question 1}} \\
\rowcolor{blue!5} - w. Using both CoTs and Explanations for SFT &87.8 &	84.2 &	79.5 &	83.8   \\
- w. Using only the Explanations for SFT & 85.3	&82.1	&78.3&	81.9 \\ 
- w. Using only the CoTs for SFT  & 81.0 & 75.8 & 68.2 & 75.0 \\ 
\midrule
\rowcolor{mypink} \multicolumn{5}{c}{\cellcolor{mypink} \textbf{Question 2}} \\
\rowcolor{blue!5} \textbf{FaithLens} ( - w. Using Correctness as Metrics) & 92.4& 	93.4 &	85.4 	&90.4 \\
- w. Using PPL as Metrics  &90.8	&91.7&	82.2&	88.2 \\
\bottomrule
\end{tabular}}
\caption{\textbf{Detailed Explainability Results for Parameter Study and Variant Methods Testing.} We use GPT-4.1 to evaluate the generated explanations from three dimensions, including readability (Read.), helpfulness (Help.), and informativeness (Info.).
}
\label{tb:para_var_exp} 
\end{table}

%% file: Tabs/para_variant_performance.tex
\begin{table*}[!t]
\centering  
\resizebox{1\hsize}{!}{
\begin{tabular}{lcccccccccccccc}
\toprule
% \toprule
 \multirow{2}{*}{\textbf{{Model}}}&

    \multirow{2}{*}{\begin{tabular}[c]{@{}c@{}}\textbf{{Agg-}}\\ \textbf{{CNN}}\end{tabular}} &

    \multirow{2}{*}{\begin{tabular}[c]{@{}c@{}}\textbf{{Agg-}}\\ \textbf{{XSum}}\end{tabular}} &

  \multirow{2}{*}{\begin{tabular}[c]{@{}c@{}}\textbf{{Claim}}\\ \textbf{{Verify}}\end{tabular}} &

  \multirow{2}{*}{\begin{tabular}[c]{@{}c@{}}\textbf{{Expert}}\\ \textbf{{QA}}\end{tabular}} &

  \multirow{2}{*}{\begin{tabular}[c]{@{}c@{}}\textbf{{FC-}}\\ \textbf{{GPT}}\end{tabular}} &

  \multirow{2}{*}{\textbf{{LfQA}}} &

  \multirow{2}{*}{\begin{tabular}[c]{@{}c@{}}\textbf{{RAG}}\\ \textbf{{Truth}}\end{tabular}} &

  \multirow{2}{*}{\textbf{{Reveal}}} &

    \multirow{2}{*}{\begin{tabular}[c]{@{}c@{}}\textbf{{Tofu-}}\\ \textbf{{MediaS}}\end{tabular}} &

    \multirow{2}{*}{\begin{tabular}[c]{@{}c@{}}\textbf{{Tofu-}}\\ \textbf{{MeetB}}\end{tabular}} &

  \multirow{2}{*}{\textbf{{Wice}}} & \multirow{2}{*}{\textbf{{HoVer}}} & \multicolumn{2}{c}{
    \begin{tabular}[c]{@{}c@{}}
        \textbf{Overall}
    \end{tabular}
}  \\
% \cmidrule(lr){14-15}
\cline{14-15}
 &  &  &  &  &  &  &  &  &   &   & & &  \textbf{Std ($\sigma$) $\downarrow$} & \textbf{Avg ($\mu$) $\uparrow$}   \\
\midrule
\rowcolor{mypink} \multicolumn{15}{c}{\cellcolor{mypink} \textbf{Study 1}} \\
\rowcolor{blue!5} \textbf{FaithLens} ($K$=10) &{84.9}	&{79.0}	&{89.4}	&{79.6}	&{92.4}	&{92.1}	&{86.8}	&{92.2}	&{85.1}	&{87.2}	&85.6	&{82.9}	&{4.6}&{86.4}   \\

- w. Setting $K$ as 6 & 84.3 	&79.1 &	88.2 &	78.3 &	90.2 &	91.1& 	86.0 &	92.2 	&84.5 &	86.2 	&84.4 	&82.6 &	4.4 &	85.6 \\
- w. Setting $K$ as 14 & 84.9 &	79.0 &	89.1 	&79.0 &	91.3 &	92.1& 	85.3 &	91.1 &	84.0 &	87.2 &	86.3 	&83.1& 	4.4 	&86.0  \\
- w. Setting $K$ as 20 & 85.2 &	78.6 &	88.2 &	80.1 &	90.2 &	92.0 &	85.3 	&90.7 	&85.5& 	88.2 	&86.3 &	82.8 &	4.1 &	86.1 \\
\midrule
\rowcolor{mypink} \multicolumn{15}{c}{\cellcolor{mypink} \textbf{Study 2}} \\
\rowcolor{blue!5} \textbf{FaithLens} (- w. Llama-Embed-Nemotron-8B) &{84.9}	&{79.0}	&{89.4}	&{79.6}	&{92.4}	&{92.1}	&{86.8}	&{92.2}	&{85.1}	&{87.2}	&85.6	&{82.9}	&{4.6}&{86.4}   \\
- w. Linq-Embed-Mistral-7B & 84.5 &	79.3 &	88.2 &	78.1 &	91.3 &	92.2 	&86.0 	&92.1 &	85.1 &	87.0 &	85.4& 	82.6 &	4.6 	&86.0  \\
- w. Gemini-Embedding-001 & 84.5 &	78.6 &	88.8 &	79.8 &	92.4 	&91.1 &	85.3 &	88.9 &	84.2 &	88.1 &	84.2 &	81.9 &	4.3 &	85.7 \\
\midrule
\rowcolor{mypink} \multicolumn{15}{c}{\cellcolor{mypink} \textbf{Study 3}} \\
\rowcolor{blue!5}  \textbf{FaithLens} (- w. Using Llama-3.1-8B-Inst) &84.9 &	79.0 &	89.4 	&79.6 &	92.4 &	92.1 	&86.8 &	92.2 &	85.1 &	87.2 &	85.6 &	82.9 &	4.6 	&86.4 \\
- w. Using Qwen-2.5-7B-Inst & 84.6	 &79.5 &	88.2 &	78.3	 &90.1 &	93.2 &	91.1	 &90.4	 &78.6 &	81.2 &	86.7 &	81.3	 &5.4 	 &85.3 \\
- w. Using DeepSeek-V3.2-Think & 84.9&	82.7	&87.1&	82.6&	90.2	&89.6&	87.1&	92.1&	79.3&	84.3&	88.3	&82.3&	3.8 &	85.9 \\
\midrule
\rowcolor{mypink} \multicolumn{15}{c}{\cellcolor{mypink} \textbf{Study 4}} \\
\rowcolor{blue!5}  \textbf{FaithLens} &84.9 &	79.0 &	89.4 	&79.6 &	92.4 &	92.1 	&86.8 &	92.2 &	85.1 &	87.2 &	85.6 &	82.9 &	4.6 	&86.4 \\
GPT-4o zero-shot &62.3&	74.9&	78.3&	68.3&	86.0	&75.0&	81.8&	86.9&	71.5	&76.9&	77.9&	73.6&	7.0	&76.1\\
GPT-4o few-shot &71.7	&76.4&	76.4&	72.7&	93.0&	90.0&	86.0&	89.0&	84.0&	85.7&	79.8&	85.4&	7.0	&82.5\\
o1 zero-shot& 68.3	&76.7&	77.1&	72.3&	85.0	&76.1	&79.6&	85.9&	65.8&	76.1&	78.7	&79.9&	5.9	&76.8 \\
o1 few-shot &82.5	&75.7&	87.5&	70.3&	93.0&	90.0	&88.0&	92.0&	81.7&	84.4&	88&	87.9&	6.7&	85.1\\
\midrule
\rowcolor{mypink} \multicolumn{15}{c}{\cellcolor{mypink} \textbf{Question 1}} \\
- w. Using both the CoTs and Explanations for SFT & 78.1& 	78.0 &	87.2& 	79.2 	&87.0 &	89.7 &	82.9 &	94.4 &	74.1 &	79.8 &	84.4 &	76.9 &	6.0 &	82.6 \\
- w. Using only the Explanations for SFT & 59.9	& 71.0	&79.8	&77.9	&85.7&	85.1&	69.1&	94.4&	69.4&	75.0	&75.4&	66.8	&9.5 &	75.8 \\ 
- w. Using only the CoTs for SFT &77.3	&74.1	&83.5	&78.7	&86.9	&89.8	&80.6	&92.2&	72.4&	78.3	&83.3	&76.8&	6.1 &	81.2 \\
\midrule
\rowcolor{mypink} \multicolumn{15}{c}{\cellcolor{mypink} \textbf{Question 2}} \\
% \rowcolor{blue!5} \textbf{FaithLens} ( - w. Using Correctness as Metrics) & 78.1& 	78.0 &	87.2& 	79.2 	&87.0 &	89.7 &	82.9 &	94.4 &	74.1 &	79.8 &	84.4 &	76.9 &	6.0 &	82.6 \\
\rowcolor{blue!5} \textbf{FaithLens} ( - w. Using Correctness as Metrics) &84.9 &	79.0 &	89.4 	&79.6 &	92.4 &	92.1 	&86.8 &	92.2 &	85.1 &	87.2 &	85.6 &	82.9 &	4.6 	&86.4 \\
- w. Using PPL as Metrics & 77.6&	78.0&	84.9	&77.9&86.9	&91.1&	83.4&	91.1&	83.2	&88.6	&84.4&	80.1	&4.9 &	83.9 \\
\bottomrule
\end{tabular}}
\caption{\textbf{Detailed Effectiveness Results for Parameter Study and Variant Methods Testing.} 
We report experimental results on 12 various datasets from LLM-AggreFact and HoVer benchmarks.}
\label{tb:para_var_per}
\end{table*}

%% file: Tabs/performance_check.tex
\begin{table*}
\centering  
\resizebox{1\hsize}{!}{
\begin{tabular}{lcccccccccccccc}
\toprule
% \toprule
 \multirow{2}{*}{\textbf{{Model}}}&

    \multirow{2}{*}{\begin{tabular}[c]{@{}c@{}}\textbf{{Agg-}}\\ \textbf{{CNN}}\end{tabular}} &

    \multirow{2}{*}{\begin{tabular}[c]{@{}c@{}}\textbf{{Agg-}}\\ \textbf{{XSum}}\end{tabular}} &

  \multirow{2}{*}{\begin{tabular}[c]{@{}c@{}}\textbf{{Claim}}\\ \textbf{{Verify}}\end{tabular}} &

  \multirow{2}{*}{\begin{tabular}[c]{@{}c@{}}\textbf{{Expert}}\\ \textbf{{QA}}\end{tabular}} &

  \multirow{2}{*}{\begin{tabular}[c]{@{}c@{}}\textbf{{FC-}}\\ \textbf{{GPT}}\end{tabular}} &

  \multirow{2}{*}{\textbf{{LfQA}}} &

  \multirow{2}{*}{\begin{tabular}[c]{@{}c@{}}\textbf{{RAG}}\\ \textbf{{Truth}}\end{tabular}} &

  \multirow{2}{*}{\textbf{{Reveal}}} &

    \multirow{2}{*}{\begin{tabular}[c]{@{}c@{}}\textbf{{Tofu-}}\\ \textbf{{MediaS}}\end{tabular}} &

    \multirow{2}{*}{\begin{tabular}[c]{@{}c@{}}\textbf{{Tofu-}}\\ \textbf{{MeetB}}\end{tabular}} &

  \multirow{2}{*}{\textbf{{Wice}}} & \multirow{2}{*}{\textbf{{HoVer}}} & \multicolumn{2}{c}{
    \begin{tabular}[c]{@{}c@{}}
        \textbf{Overall}
    \end{tabular}
}  \\
% \cmidrule(lr){14-15}
\cline{14-15}
 &  &  &  &  &  &  &  &  &   &   & & &  \textbf{Std ($\sigma$) $\downarrow$} & \textbf{Avg ($\mu$) $\uparrow$}   \\
\midrule
\rowcolor{mypink} \multicolumn{15}{c}{\cellcolor{mypink} \textbf{Results by using the modified prompt shown in Figure \ref{fig:prompt_to_generate_exp} for baselines}} \\
GPT-4o & 63.2 &	73.7 &	78.3 &	69.2 &	84.8 	&75.0 &	81.2 &	86.4 &	70.5 &	75.3 &	76.3 	&74.6& 	6.5 &	75.7  \\
o1  &67.3 &	75.6 &	78.3 	&73.2 &	84.8& 	76.5 &	80.1 	&84.3 	&66.7 &	76.5 &	79.2 &	80.8 &	5.7 &	76.9  \\
DeepSeek-V3.2-Non-Think & 63.4 &	66.3 &	78.0 	&78.0 &	89.0 &	70.0 &	81.9 	&91.0 &	69.6 &	84.0 	&71.9& 	74.8 &	8.8 &	76.5  \\
DeepSeek-V3.2-Think & 84.3& 	73.5 &	87.8 &	76.8 &	89.1 &	77.1 	&86.8 &	91.0 &	77.5 &	85.2 &	83.0 &	74.8 &	6.0 &	82.2 \\
Claude-3.7-Sonnet & 77.3 &	75.2& 	82.6 &	75.1 &	87.0 &	88.9 &	88.9& 	87.3 &	85.1 &	84.0 &	87.5 &	81.2 &	5.1 &	83.3  \\
Llama-3.1-405B-Inst &67.3	&72.1	&81.3	&65.6&	80.9&	77.8	&82.3&	86.2	&65.2&	80.1	&73.2&	81.6&	7.2 	&76.1   \\
Llama-3.1-8B-Inst &42.0 	&47.2 &	62.6 &	48.1 &	70.1& 	49.1 &	52.4 &	77.2 	&51.3& 	64.2 &	52.1 &	65.2 	&10.7 &	56.8  \\
GPT-4.1 & 74.1 &	72.5 &	88.0 &	78.4 	&92.4 &	84.9 &	88.0 	&92.1 &	80.4 &	90.1 &	84.3 &	82.8 &	6.6 &	84.0 \\
o3-mini & 65.3 &	82.1 	&81.3 &	74.2 	&85.6 &	82.0 &	81.0& 	85.3 &	77.6& 	79.6 &	80.9 	&80.1 &	5.4 &	79.6  \\
o3 & 68.1 &	80.1 &	82.6 &	80.5 &	87.0 &	88.9 &	78.2 &	92.2 &	83.6 &	84.1 &	83.3 &	81.5 	&6.0 &	82.5  \\
GPT-5.2 & 88.2 &	72.1& 	88.8 &	85.3 &	90.1 &	88.6 &	88.8 &	92.7 &	87.6 &	80.1 &	88.6 &	81.6 &	5.6 &	86.0 \\
\bottomrule
\end{tabular}}
\caption{\textbf{Effectiveness Results by Using the Modified Prompt Shown in Figure \ref{fig:prompt_to_generate_exp} for Baselines to Generate the Corresponding Explanations.}
Combined with the results from Table \ref{tb:main}, we cannote that requiring an explanation before the final answer for LLM-based baselines has little to no effect on the numeric prediction outcome.
}
\label{tb:effectiveness_check}
\end{table*}

%% file: Tabs/prompt_training.tex
\begin{figure*}[t]
\centering
\begin{tcolorbox}[title=Prompt used for training and inference of FaithLens, width=\textwidth, colback=gray!5, colframe=black, fonttitle=\bfseries]
\small
Determine whether the provided claim is consistent with the corresponding document. 

Consistency in this context implies that all information presented in the claim is substantiated by the document. If not, it should be considered inconsistent. \\

- First, think step by step about whether all the information in the claim is fully supported by the document within <think> and </think> tags. \\

- Then, please provide an easy-to-understand explanation for your answer within <reason> and </reason> tags. \\

- Finally, assess the claim's consistency with the document by responding with either ``Yes'' or ``No'' and wrap your final answer in <answer> and </answer> tags. \\

Document: [DOCUMENT]

Claim: [CLAIM]

\end{tcolorbox}
\caption{Prompt used for training and inference of FaithLens.}
\label{fig:prompt_train_inference}
\end{figure*}

\begin{figure*}[t]
\centering
\begin{tcolorbox}[title=Prompt used for data synthesis, width=\textwidth, colback=gray!5, colframe=black, fonttitle=\bfseries]
\small
Determine whether the provided claim is consistent with the corresponding document. 

Consistency in this context implies that all information presented in the claim is substantiated by the document. If not, it should be considered inconsistent. \\

- First, think step by step about whether all the information in the claim is fully supported by the document within <think> and </think> tags. \\

- Then, please provide an easy-to-understand explanation for your answer within <reason> and </reason> tags. \\

- Finally, assess the claim's consistency with the document by responding with either ``Yes'' or ``No'' and wrap your final answer in <answer> and </answer> tags. \\

Document: [DOCUMENT]

Claim: [CLAIM]

\end{tcolorbox}
\caption{Prompt used for data synthesis.}
\label{fig:prompt_data_syn}
\end{figure*}

\begin{figure*}[t]
\centering
\begin{tcolorbox}[title=Prompts used for our designed explanation quality filtering, width=\textwidth, colback=gray!5, colframe=black, fonttitle=\bfseries]
\small
Determine whether the provided claim is consistent with the corresponding document. 

Consistency in this context implies that all information presented in the claim is substantiated by the document. If not, it should be considered inconsistent. \\

- First, think step by step about whether all the information in the claim is fully supported by the document within <think> and </think> tags. \\

- Finally, assess the claim's consistency with the document by responding with either ``Yes'' or ``No'' and wrap your final answer in <answer> and </answer> tags. \\

Document: [DOCUMENT]

Claim: [CLAIM] \\

\textcolor{purple}{<think>[CoT]</think><answer>[Answer]</answer>} \\

- - -\\

Determine whether the provided claim is consistent with the corresponding document. 

Consistency in this context implies that all information presented in the claim is substantiated by the document. If not, it should be considered inconsistent. \\

- First, think step by step about whether all the information in the claim is fully supported by the document within <think> and </think> tags. \\

- Then, please provide an easy-to-understand explanation for your answer within <reason> and </reason> tags. \\

- Finally, assess the claim's consistency with the document by responding with either ``Yes'' or ``No'' and wrap your final answer in <answer> and </answer> tags. \\

Document: [DOCUMENT]

Claim: [CLAIM] \\

\textcolor{purple}{<think>[CoT]</think><reason>[Explanation]</reason><answer>[Answer]</answer>}

\end{tcolorbox}
\caption{Prompts used for our designed explanation quality filtering.
To assess the explanation quality, we concatenate the generated CoT (<think>) and explanation (<reason>) as the input and compute the perplexity of the corresponding [answer]. 
The upper part of the prompt is used to measure the model’s perplexity for the ground-truth label using only the document $doc$, claim $c$, and the synthetic CoT $\hat{cot}$. 
The lower part of the prompt reflects the model’s confidence in generating the correct label by conditioning on the tested explanation, i.e., by concatenating both the synthetic CoT and the corresponding explanation as inputs.
}
\label{fig:prompt_exp_filtering}
\end{figure*}

\begin{figure*}[t]
\centering
\begin{tcolorbox}[title=Prompts used for our designed data diversity filtering, width=\textwidth, colback=gray!5, colframe=black, fonttitle=\bfseries]
\small
Determine whether the provided claim is consistent with the corresponding document. 

Consistency in this context implies that all information presented in the claim is substantiated by the document. If not, it should be considered inconsistent. \\

- First, think step by step about whether all the information in the claim is fully supported by the document within <think> and </think> tags. \\

- Then, please provide an easy-to-understand explanation for your answer within <reason> and </reason> tags. \\

- Finally, assess the claim's consistency with the document by responding with either ``Yes'' or ``No'' and wrap your final answer in <answer> and </answer> tags. \\

Document: [DOCUMENT]

Claim: [CLAIM] \\

\textcolor{purple}{<think>[CoT]</think><reason>[Explanation]</reason><answer>[Answer]</answer>} \\

- - -\\

Determine whether the provided claim is consistent with the corresponding document. 

Consistency in this context implies that all information presented in the claim is substantiated by the document. If not, it should be considered inconsistent. \\

- First, think step by step about whether all the information in the claim is fully supported by the document within <think> and </think> tags. \\

- Then, please provide an easy-to-understand explanation for your answer within <reason> and </reason> tags. \\

- Finally, assess the claim's consistency with the document by responding with either ``Yes'' or ``No'' and wrap your final answer in <answer> and </answer> tags. \\

Document: [DOCUMENT]

Claim: [CLAIM] \\

Example: [Tested Sample] \\

\textcolor{purple}{<think>[CoT]</think><reason>[Explanation]</reason><answer>[Answer]</answer>}

\end{tcolorbox}
\caption{Prompts used for data diversity filtering.
The upper part of the prompt is used to measure the model’s perplexity for the ground-truth label based the document $doc'$, claim $c'$, the synthetic CoT $\hat{cot}'$ and explanation $\hat{e}'$. 
The lower part of the prompt reflects the model’s confidence in generating the correct label by incorporating candidate sample $\hat{s}$ as an in-context demonstration and recompute the perplexity.
}
\label{fig:prompt_diversity_filtering}
\end{figure*}

\begin{figure*}[t]
\centering
\begin{tcolorbox}[title=Prompt used for computing explanation quality reward, width=\textwidth, colback=gray!5, colframe=black, fonttitle=\bfseries]
\small
Determine whether the provided claim is consistent with the corresponding document. 

Consistency in this context implies that all information presented in the claim is substantiated by the document. If not, it should be considered inconsistent. \\

- First, please refer to the provided explanation to assist you to answer the question. \\

- Then, please assess the claim's consistency with the document by responding with either ``Yes'' or ``No''. Please wrap your final answer in <answer> and </answer>. \\

Document: [DOCUMENT]

Claim: [CLAIM]

Explanation: [EXPLANATION]

\end{tcolorbox}
\caption{Prompt used for computing explanation quality reward.
}
\label{fig:prompt_exp_reward}
\end{figure*}

\begin{figure*}[t]
\centering
\begin{tcolorbox}[title=Prompt used for scoring the generated explanations using the LLM as a judge, width=\textwidth, colback=gray!5, colframe=black, fonttitle=\bfseries]
\small
You are an evaluator. Another model was tasked with assessing whether a source document supports a given claim, and it successfully arrived at the correct determination based on the provided task instruction. 
The model then generated an explanation for its conclusion. 
Your role is to evaluate the quality of that explanation along the specified dimensions. \\

\#\#\#\ Scoring Criteria:

1. Readability (1–5): The explanation should be written in a clear and well-structured manner that enables the reader to easily follow the reasoning behind the model’s conclusion. Beyond sentence fluency, focus on whether the explanation presents ideas in a logical sequence, avoids ambiguity, and makes it straightforward for the user to correctly understand why the model arrived at its prediction.

2. Helpfulness (1–5): The explanation should effectively guide the user to understand why the model arrived at its conclusion. Focus on whether the reasoning is clear and logically connected to the claim and document, enabling the user to act on, adapt, or reconsider the claim if needed.

3. Informativeness (1–5): The explanation should provide detailed, specific, and substantive information relevant to the claim and document. Focus on the richness and completeness of content, such as explicit evidence cited, nuanced reasoning, or contextual details that give a deeper understanding, even beyond what is strictly needed to justify the conclusion. \\

\#\#\# Output Format (JSON only):

\{

  ``readability'': <1-5>,
  
  ``helpfulness'': <1-5>,
  
  ``informativeness'': <1-5>
  
\} \\

\#\#\# Task Instruction (includes the claim and document):

[Task Instruction] \\ 

\#\#\# Explanation to Evaluate:

[Explanation\_Text] 

\end{tcolorbox}
\caption{Prompt used for scoring the generated explanations using the LLM as a judge.
}
\label{fig:llm-as-a-judge}
\end{figure*}

\begin{figure*}[t]
\centering
\begin{tcolorbox}[title=Prompts used for evaluating LLM-based baselines, width=\textwidth, colback=gray!5, colframe=black, fonttitle=\bfseries]
\small
Instructions:

1. You have been given a STATEMENT and some DOCUMENT.

2. Determine whether the given STATEMENT is supported by the given DOCUMENT. The STATEMENT does not need to be explicitly supported by the DOCUMENT but should be strongly implied by the DOCUMENT.

3. Before showing your answer, think step-by-step and show your specific reasoning. As part of your reasoning, summarize the main points of the DOCUMENT.

4. If the STATEMENT is supported by the DOCUMENT, be sure to show the supporting evidence.

5. After stating your reasoning, restate the STATEMENT and then determine your final answer based on your reasoning and the STATEMENT.

6. Your final answer should be either [Attributable] or [Not Attributable], or [Contradictory].

7. Wrap your final answer in square brackets. \\

DOCUMENT:

[DOCUMENT PLACEHOLDER]

STATEMENT:

[STATEMENT PLACEHOLDER] \\

- - -\\

Instructions:

1. You have been given a STATEMENT and some DOCUMENT.

2. Determine whether the given STATEMENT is supported by the given DOCUMENT. The STATEMENT does not need to be explicitly supported by the DOCUMENT, but should be strongly implied by the DOCUMENT.

3. Before showing your explanation and answer, think step-by-step and show your chain of thought and specific reasoning. As part of your reasoning, summarize the main points of the DOCUMENT.

4. If the STATEMENT is supported by the DOCUMENT, be sure to show the supporting evidence.

5. After stating your reasoning, restate the STATEMENT and then determine your final answer based on your reasoning and the STATEMENT.

6. After your reasoning but before the final answer, provide a human-readable explanation (<explanation>) that clearly and concisely justifies your conclusion, citing specific parts or descriptions from the DOCUMENT that support or contradict the STATEMENT. This explanation should be understandable to a human reader and should not reveal the model's internal chain of thought.

7. Your final answer should be either [Attributable] or [Not Attributable], or [Contradictory].Wrap your final answer in square brackets.

8. Your final output must follow the exact structure:
   <think>step-by-step reasoning (your internal reasoning)</think>
   <reason>human-readable justification using evidence from the document</reason>
   <answer>[Attributable] or [Not Attributable] or [Contradictory]</answer> \\

DOCUMENT:

[DOCUMENT PLACEHOLDER]

STATEMENT:

[STATEMENT PLACEHOLDER]
\end{tcolorbox}
\caption{Prompts used for evaluating LLM-based baselines.
The upper part of prompt is adapted from \citet{seo2025verifying} and is used to evaluate the effectiveness of LLM-based baselines. 
The label ``Attributable'' is mapped to the absence of hallucination, while the labels ``Not Attributable'' and ``Contradictory'' are mapped to the presence of hallucination. 
The prompt below shows our modification to the original prompt, enabling the model to output both an explanation and a final prediction, without affecting the model's final prediction performance.
}
\label{fig:prompt_to_generate_exp}
\end{figure*}

\begin{figure*}[t]
\centering
\begin{tcolorbox}[title=Prompt used for claim decontextualization, width=\textwidth, colback=gray!5, colframe=black, fonttitle=\bfseries]
\small
You are provied with a context and a claim. Please first determine if the claim can stand alone whitout the conext. If not, provide a decontextualzied version of the claim that incorporates necessary information from the context to make it self-contained.
The revision should be as minimum as possible. Please respond with a JSON format: \{``label'': ``yes''/``no'', ``decontext'':
``NA''/decontextualized claim\}. \\

Example 1:

Context: There are many reasons why poetry is important for children. Poetry can help children build confidence through memorizing and reciting poems. It can also provide an easy way for children to remember a lesson or value.

Claim: It can also provide an easy way for children to remember a lesson or value.

Answer: \{``label'': ``no'', ``decontext'': ``Poetry can provide an easy way for children to remember a lesson or value.''\} \\

Example 2:
Context: Yes, ancient societies had concepts of rights. The concept of rights first appeared in the theory of natural law which existed in the state of nature. In this state, people enjoyed certain rights sanctioned by natural law.

Claim: In this state, people enjoyed certain rights sanctioned by natural law.

Answer: \{``label'': ``no'', ``decontext'': ``In the state of nature, people enjoyed certain rights sanctioned by natural law''\} \\

Example 3:

Context: The ancient Greeks had some concept of human rights, although there is no single word in classical Greek that captures the sense of ``rights'' as it is used in modern political thought. However, Greek customs and institutions provided protection to private property unique in the ancient world, instilling a strong sense of equality. The idea of human rights spread quickly from Babylon to Greece and eventually Rome, where the concept of ``natural law'' arose.

Claim: The idea of human rights spread quickly from Babylon to Greece and eventually Rome, where the concept of ``natural law'' arose.

Answer: \{``label'': ``yes'', ``decontext'': ``NA''\} \\

Your Turn:

Context: [CONTEXT]

Claim: [CLAIM]

Answer:
\end{tcolorbox}
\caption{Prompt used for claim decontextualization.
}
\label{fig:claim_1}
\end{figure*}

\begin{figure*}[t]
\centering
\begin{tcolorbox}[title=Prompt used for claim decomposition, width=\textwidth, colback=gray!5, colframe=black, fonttitle=\bfseries]
\small
Segment the following sentence into individual facts:\\

Sentence: Other title changes included Lord Steven Regal and The Nasty Boys winning the World Television Championship and the World Tag Team Championship respectively.

Facts:

- Lord Steven Regal won the World Television Championship.

- The Nasty Boys won the World Tag Team Championship. \\

Sentence: The parkway was opened in 2001 after just under a year of construction and almost two decades of community requests.

Facts:

- The parkway was opened in 2001.

- The parkway was opened after just under a year of construction.

- The parkway was opened after two decades of community requests. \\

Sentence: Touring began in Europe in April-June with guitarist Paul Gilbert as the opening act, followed by Australia and New Zealand in July, Mexico and South America in late July-August, and concluding in North America in October-November.

Facts:

- Touring began in Europe in April-June.

- The opening act of the tour was guitarist Paul Gilbert.

- The tour was in Australia and New Zealand in July.

- The tour was in Mexico and South America in late July-August.

- The tour was concluded in North America in October-November. \\

Sentence: In March 2018, the company partnered With Amazon Web Services (AWS) to offer Al-enabled conversational solutions to customers in India.

Facts:

- The company partnered with Amazon Web Services (AWS) in March 2018.

- The two companies partnered to offer Al-enabled conversational solutions to customers in India. \\

Sentence: The most significant of these is in Germany, which now has a Yazidi community of more than 200,000 living primarily in Hannover, Bielefeld, Celle, Bremen, Bad Oeynhausen, Pforzheim and Oldenburg.

Facts:

- The most significant of these is in Germany.

- Germany now has a Yazidi community of more than 200,000.

- Yazidi community in Germany lives primarily in Hannover, Bielefeld, Celle, Bremen, Bad Oeynhausen, Pforzheim and Oldenburg. \\

Sentence: A previous six-time winner of the Nations’ Cup, Sebastian Vettel became Champion of Champions for the first time, defeating Tom Kristensen, who made the final for the fourth time, 2-0.

Facts:

- Sebastian Vettel is a previous six-time winner of the Nations’ Cup.

- Sebastian Vettel became Champion of Champions for the first time, defeating Tom Kristensen, 2-0.

- Tom Kristensen made the final for the fourth time.  \\

Sentence: [SENTENCE]

Facts:

\end{tcolorbox}
\caption{Prompt used for claim decomposition.
}
\label{fig:claim_2}
\end{figure*}

\begin{figure*}[t]
\centering
\begin{tcolorbox}[title=The principles of human evaluation, width=\textwidth, colback=gray!5, colframe=black, fonttitle=\bfseries]
\small
You are asked to evaluate the responses generated by different models. You should choose the preferred responses according to the following perspectives independently: \\

1. Readability: The explanation should be written in a clear and well-structured manner that enables the reader to easily follow the reasoning behind the model’s conclusion. Beyond sentence fluency, focus on whether the explanation presents ideas in a logical sequence, avoids ambiguity, and makes it straightforward for the user to correctly understand why the model arrived at its prediction.

2. Helpfulness: The explanation should effectively guide the user to understand why the model arrived at its conclusion. Focus on whether the reasoning is clear and logically connected to the claim and document, enabling the user to act on, adapt, or reconsider the claim if needed.

3. Informativeness: The explanation should provide detailed, specific, and substantive information relevant to the claim and document. Focus on the richness and completeness of content, such as explicit evidence cited, nuanced reasoning, or contextual details that give a deeper understanding, even beyond what is strictly needed to justify the conclusion. \\

Finally, please make a decision among the 3 opinions, including FaithLens Wins, Tie, and GPT-4o Wins.

\end{tcolorbox}
\caption{The principles of human evaluation.
}
\label{fig:prompt_human_eval}
\end{figure*}

\begin{figure*}[t]
\centering
\begin{tcolorbox}[title=Prompt used for question 1 in variant methods testing, width=\textwidth, colback=gray!5, colframe=black, fonttitle=\bfseries]
\small
Determine whether the provided claim is consistent with the corresponding document. 

Consistency in this context implies that all information presented in the claim is substantiated by the document. If not, it should be considered inconsistent. \\

- First, please provide an easy-to-understand explanation for your answer within <reason> and </reason> tags. \\

- Finally, assess the claim's consistency with the document by responding with either ``Yes'' or ``No'' and wrap your final answer in <answer> and </answer> tags. \\

Document: [DOCUMENT]

Claim: [CLAIM]

\end{tcolorbox}
\caption{Prompt used for question 1 in variant methods testing.}
\label{fig:only_exp}
\end{figure*}

\begin{figure*}[t]
\centering
\begin{tcolorbox}[title=Prompt used for question 1 in variant methods testing, width=\textwidth, colback=gray!5, colframe=black, fonttitle=\bfseries]
\small
Determine whether the provided claim is consistent with the corresponding document. 

Consistency in this context implies that all information presented in the claim is substantiated by the document. If not, it should be considered inconsistent. \\

- First, think step by step about whether all the information in the claim is fully supported by the document within <think> and </think> tags. \\

- Finally, assess the claim's consistency with the document by responding with either ``Yes'' or ``No'' and wrap your final answer in <answer> and </answer> tags. \\

Document: [DOCUMENT]

Claim: [CLAIM]

\end{tcolorbox}
\caption{Prompt used for question 1 in variant methods testing.}
\label{fig:only_cot}
\end{figure*}

\begin{figure*}[t]
\centering
\begin{tcolorbox}[title=Case study from LLM-AggreFact, width=\textwidth, colback=gray!5, colframe=black, fonttitle=\bfseries]
\small
\textbf{Document}: \{Document Content, approximately 3,500 words\} \\
\textbf{Claim}: The federal Lanham Act and the FTC Act both prohibit unfair or deceptive advertising. \\

\textbf{GPT-4o}:
The document details the FTC Act’s provisions against unfair or deceptive advertising practices, explaining the requirements for truthfulness and evidence-based substantiation of claims. 
It highlights the importance of not misleading consumers and presents various FTC guidelines and rules in support of this.
However, it does not mention the Lanham Act, and thus there is nocontent in the DOCUMENT that supports the claim regarding the Lanham Act's role in prohibiting deceptive advertising. \\

\textbf{o1}: The DOCUMENT references the FTC Act’s prohibition on unfair or deceptive advertising but does not mention the Lanham Act at all.
Consequently, there is no information confirming that the Lanham Act prohibits such advertising in the DOCUMENT. \\

\includegraphics[height=1.3em]{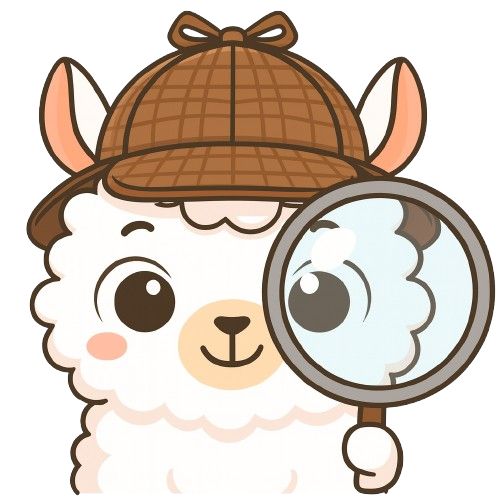} \textbf{FaithLens (Ours)}: The claim states that both the federal Lanham Act and the FTC Act prohibit unfair or deceptive advertising. 
The document explicitly discusses the FTC Act, which it mentions prohibits ``unfair or deceptive advertising in any medium.'' 
This directly supports the claim regarding the FTC Act. 
However, the document does not mention the Lanham Act at all. it lists other federal laws like the truth in lending act, fair credit billing act, fair credit reporting act, equal credit opportunity act, and others, but no reference is made to the Lanham Act. 
Since the claim asserts that the Lanham Act also prohibits unfair or deceptive advertising, and this information is not present in the document, the claim is not fully substantiated. 
For the claim to be consistent, all information in it must be supported by the document, which is not the case here.

\end{tcolorbox}
\caption{Case study from LLM-AggreFact.}
\label{fig:case_aggre}
\end{figure*}

\begin{figure*}[t]
\centering
\begin{tcolorbox}[title=Case study from HoVer, width=\textwidth, colback=gray!5, colframe=black, fonttitle=\bfseries]
\small
\textbf{Document}: Animation is the process of making the illusion of motion and the illusion of change by means of the rapid succession of sequential images that minimally differ from each other. 
The illusion—as in motion pictures in general—is thought to rely on the phi phenomenon and beta movement, but the exact causes are still unclear. 
Tom and Jerry: A Nutcracker Tale is a 2007 animated holiday themed fantasy comedy direct-to-video film starring ``Tom and Jerry'' produced by Warner Bros. 
Animation and directed by Spike Brandt and Tony Cervone. 
It is loosely based on the story ``The Nutcracker and the Mouse King'' by E. T. A. Hoffmann, with Jerry in the role of the Nutcracker, while Tom played one of the King Mouse's (who, in this version, is replaced with a cat) henchmen. \\
\textbf{Claim}: Tom and Jerry: A Nutcracker Tale, released in 1940, is an example of an Animation motion picture that makes the illusion of motion and change by the rapid succession of sequential images that minimally differ from each other.\\

\textbf{GPT-4o}: According to the DOCUMENT, ``Tom and Jerry: A Nutcracker Tale'' is indeed an animated film that employs the technique described—creating the illusion of motion through successive images that minimally differ. 
However, the STATEMENT incorrectly lists its release year as 1940, whereas the DOCUMENT states it was released in 2007. 
The incorrect release year makes the STATEMENT contradictory to the DOCUMENT. \\

\textbf{o1}: The DOCUMENT states that Tom and Jerry: A Nutcracker Tale is a 2007 animated film, which conflicts with the STATEMENT’s assertion that it was released in 1940. 
While it is indeed an example of an animated motion picture (as described in the DOCUMENT’s explanation of animation), the mismatch in release date renders the STATEMENT contradictory overall. \\

\includegraphics[height=1.3em]{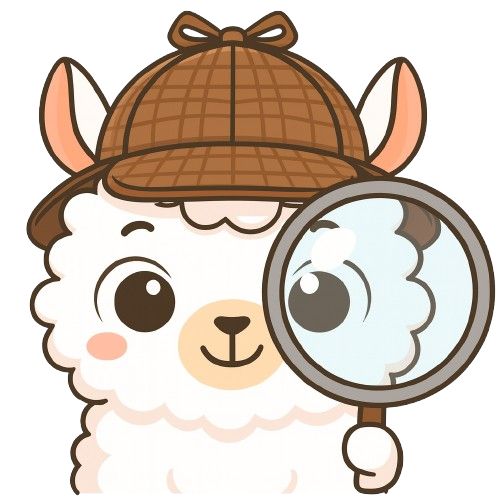} \textbf{FaithLens (Ours)}: The claim states that ``Tom and Jerry: A Nutcracker tal'' was released in 1940. 
However, the document explicitly says the film is a ``2007 animated holiday themed fantasy comedy.'' 
Since the release year (2007) is directly provided in the document, the claim's assertion that it was released in 1940 is not supported.
Additionally, the claim calls the film an ``Animation motion picture'' and describes 
it as making motion through ``rapid succession of sequential images,'' which aligns with the document's description of animation. 
Since the year 1940 is incorrect based on the document, the claim contains information not substantiated by the document.

\end{tcolorbox}
\caption{Case study from HoVer.}
\label{fig:case_hover}
\end{figure*}